\documentclass{article}
\usepackage{arxiv}
\usepackage[utf8]{inputenc} 
\usepackage[T1]{fontenc}    
\usepackage{hyperref}       
\usepackage{url}            
\usepackage{booktabs}       
\usepackage{amsfonts}       
\usepackage{nicefrac}       
\usepackage{microtype}      
\usepackage{lipsum}
\usepackage{graphicx}
\usepackage{amsmath}
\usepackage{breqn}
\usepackage[thinlines]{easytable}
\usepackage{array}
\usepackage{multirow}
\usepackage{caption} 
\graphicspath{ {./images/} }

\title{DACB-Net: Dual Attention Guided Compact Bilinear Convolution Neural Network for Skin Disease Classification}

\author{
 Belal Ahmad \\
  School of Computer Science and Technology, \\
  Huazhong University of Science and Technology, \\
  Wuhan, China \\
  \texttt{ahmadbelal@hust.edu.cn} \\
   \And
 Mohd Usama \\
  Departments of Diagnostics and Intervention, \\
  and Biomedical Engineering,\\
  Umea University, Sweden \\
  \texttt{mohd.usama@umu.se} \\
  \And
 Tanvir Ahmad \\
  School of Information and Control Engineering, \\
  Xi’an University of Architecture and Technology, \\
  Xian, China \\
  \texttt{tanvir@xauat.edu.cn} \\
  \And
 Adnan Saeed \\
  School of Civil and Hydraulic Engineering, \\
  Huazhong University of Science and Technology, \\
  Wuhan, China \\
  \texttt{adnansaeed@hust.edu.cnn} \\
  \And
 Shabnam Khatoon \\
  School of Management Science and Engineering,\\
  China University of Geosciences, \\
  Wuhan, China \\
  \texttt{shabnamali.ali9@gmail.com} \\
  \And
  Min Chen\\
  School of Computer Science and Technology, \\
  Huazhong University of Science and Technology, \\
  Wuhan, China \\
    \texttt{minchen2O12@hust.edu.cn} \\
}

\begin{document}
\maketitle
\begin{abstract}
This paper introduces the three-branch Dual Attention-Guided Compact Bilinear CNN (DACB-Net) by focusing on learning from disease-specific regions to enhance accuracy and alignment. A global branch compensates for lost discriminative features, generating Attention Heat Maps (AHM) for relevant cropped regions. Finally, the last pooling layers of global and local branches are concatenated for fine-tuning, which offers a comprehensive solution to the challenges posed by skin disease diagnosis. Although current CNNs employ Stochastic Gradient Descent (SGD) for discriminative feature learning, using distinct pairs of local image patches to compute gradients and incorporating a modulation factor in the loss for focusing on complex data during training. However, this approach can lead to dataset imbalance, weight adjustments, and vulnerability to overfitting. The proposed solution combines two supervision branches and a novel loss function to address these issues, enhancing performance and interpretability. The framework integrates data augmentation, transfer learning, and fine-tuning to tackle data imbalance to improve classification performance, and reduce computational costs. Simulations on the HAM10000 and ISIC2019 datasets demonstrate the effectiveness of this approach, showcasing a 2.59\% increase in accuracy compared to the state-of-the-art.
\end{abstract}

\keywords{Skin Disease Classification \and Bilinear Convolutional Neural Networks \and Attention Heat Maps \and Stochastic Gradient Descent.}

\section{Introduction}
\paragraph{} Skin lesions are deviations in the dermal structure that show abnormal growth or appearance compared to the surrounding skin area. They are categorized as either primary or secondary lesions based on their origin. Primary lesions manifest as abnormal skin conditions present at birth, acquired during one’s lifetime, or at the onset of an illness \cite{Russakovsky2015}. On the other hand, secondary lesions result from modifications to a primary lesion caused by treatment, disease progression, or manual intervention such as scratching, pinching, or rubbing \cite{Young2004}. Depending on the type of skin lesion, dermatologists recommend various treatments, including surgery, medications, or home care. Despite their seemingly innocuous nature, some skin lesions may pose hazards, signalling the potential presence of skin cancer and necessitating surgical removal. Melanoma, in particular, stands out as the riskiest form of skin cancer. While deadly when it spreads, melanoma is curable in its early stages. Hence, accurate diagnosis of skin lesions is crucial to ensure timely treatment.
\paragraph{} Globally, approximately 123,000 melanoma cases and 3,000,000 non-melanoma cases are reported annually. Factors like ultraviolet radiation and sun overexposure are the main reasons for an annual increase in skin cancer cases by 3\% to 7\%, translating to an additional 4500 melanoma cases \cite{Li2014}. A study revealed that dermatologists with over ten years of experience achieved an 80\% accuracy in skin lesion diagnosis, while those with 3--5 years of experience correctly diagnosed 62\% of cases \cite{Chaturvedi2020}. Various techniques are employed to identify potentially dangerous lesions, including dermoscopy, complete physical exams, biopsies, medical history analysis, skin exams, and monitoring changes in lesion time, size, shape, and associated symptoms. Despite stabilized mortality rates, skin cancer still imposes significant morbidity and treatment costs \cite{Inthiyaz2023}.
\paragraph{} To address these challenges, developing a Computer-Aided Diagnosis (CAD) system becomes imperative, utilizing a unified measurement standard to assess lesion severity. In recent years, integrating artificial intelligence and computer technology has played a pivotal role in various disciplines. Medical image processing, a multidisciplinary subject bridging medicine and computer technology, has leveraged image processing technology, computer vision, and other methods to standardize the analysis of medical images, contributing to unified diagnostic results. DL-based classification methods have gained prominence in the classification of skin diseases, offering a potential solution to the complexities associated with skin lesion diagnosis. In recent years, CNNs have played a crucial role in object classification \cite{Sushma2024} and feature learning \cite{Guannan2024}, significantly improving the performance of natural image classification \cite{Inthiyaz2023} and skin disease classification \cite{Moeskops2016}. Nevertheless, current methods often depend solely on global features extracted by the FC layer, overlooking valuable local pattern features found in the intermediate layers of Deep CNN (DCNN). Because of the evident intra-class variation and inter-class similarity, the subtle differences in local patterns in the skin lesion regions are critical for skin disease classification \cite{Codella2015} \cite{Demyanov2016}.
\paragraph{} Traditionally, two-stage approaches have been employed to enhance the extraction of local skin lesion features. For instance, Moeskops et al. \cite{Moeskops2016} introduced an additional segmentation mask to train a segmentation network, which was utilized to crop skin lesion regions from the entire image. Subsequently, the classification network focused on the cropped images during training, improving performance by emphasizing skin lesion region features. However, this method faced limitations, such as the scarcity and expense of segmentation mask annotations, demanding considerable professional knowledge and labor. In another approach, Yu et al. \cite{Alzubaidi2021} utilized the Fisher-Vector (FV) encoding method to aggregate DCNN features from re-scaled input images. The FV encoding features were then employed to train a chi-squared kernel-based SVM for melanoma recognition. Yan et al \cite{Yan2019} proposed an end-to-end network for melanoma recognition. They incorporated an attention module to focus on skin lesion local features within intermediate layers \cite{Schlemper2018}. They employed an additional segmentation mask as prior information to refine the attention map decisively. However, using the segmentation mask for the entire skin lesion region in the segmentation task while introducing intense supervision proved effective in refining local pattern features due to significant variations in scale, shape, and site across different stages and degrees of skin cancer. Building on insights from Yosinski et al. \cite{Yosinski2015}, who demonstrated that different feature channels of extracted CNN features can represent various local pattern features in the original input image; a recent development is GCNet \cite{Yue2019}. GCNet introduced a lightweight non-local block to capture the non-local-based global context in the feature map, facilitating the capture of long-distance dependencies between pixels. Consequently, this approach highlights discriminative local pattern features, addressing the limitations of traditional convolution layers with limited receptive fields, especially in capturing the evolving appearance of local patterns within changing skin lesion regions.
\paragraph{} To address the challenges, this paper introduces a novel Dual Attention Mechanism (DAM)-based network with a novel loss function that combines complementary entropy and cross-entropy for improved skin disease classification. The novel DAM aims to capture meaningful local patterns within the skin lesion region. In detail, the DAM spatially focuses on the multi-scale skin lesion region features present in the intermediate layers of the backbone network. Importantly, it achieves this without requiring complicated pre-processing and automatically reduces irrelevant artifact features. Given the substantial changes in skin lesions, the spatially filtered multi-scale skin lesion region features are concatenated to form the output feature. To further refine this output feature and extract meaningful local pattern features within the skin lesion regions, the proposed approach deviates from using an extra segmentation mask as intense supervision \cite{Yan2019}. Inspired by the research \cite{Yosinski2015} \cite{Yue2019}, a Channel Attention (CA) module is introduced. This module employs a lightweight non-local block to extract the non-local-based global feature of the Spatial Attention (SA). The SA captures long-distance dependencies between each pair of pixels in the SA output feature (SA\_O). Subsequently, the non-local-based global feature is utilized to generate a feature channel reweighting vector. This vector is then applied to reweight the different feature channels of the SA\_O, forming the output feature of the CA\_O.
\paragraph{} Consequently, the CA\_O automatically highlights meaningful local pattern features critical for skin disease classification. This not only enhances the interpretability of the proposed network but also contributes to the efficient utilization of small training data and optimization of the entire network. Integrating the DAM and Channel Attention Module (CAM) addresses the challenges posed by significant changes in skin lesions, providing a robust framework for accurate and interpretable skin disease classification.
\paragraph{} The proposed network combines two supervision branches and a new loss function: CCE. The two branches employ the CA\_O and the global feature captured by the last convolutional layer as input, respectively. The image-level labels for classification supervise both of these two branches. SGD is mainly used in the discriminative feature learning for skin lesions, which computes the gradient from distinct pairs of local image patches and the modulation factor in the loss to focus the model on the training of complex samples that also affect the imbalance of the dataset by adjusting the weight. As a result, it fails to address the issue of achieving high accuracy in both simple and complex models and may lead to vulnerability to overfitting.
\paragraph{} This paper introduces a novel loss function based on complementary and cross-entropy to address the earlier challenges. By leveraging training samples from non-labeled categories, the method harnesses incorrect class information to train a robust classification model, particularly adept at handling unbalanced class distributions. This approach effectively mitigates the impact of uneven data distribution on the network model’s predictive ability without increasing the number of samples for specific categories. Consequently, the strategy enhances learning opportunities for underrepresented sample types, compelling the network model to prioritize diseased samples and improving overall classification ability. Extensive experiments are conducted on the HAM10000 \cite{Tschandl2019} and ISIC2019 \cite{Kassem2020} datasets to evaluate the performance of the proposed network, demonstrating state-of-the-art results across both datasets. The primary contributions of the proposed method are summarized as follows:
\begin{itemize}
\item
A novel DAM is proposed to highlight meaningful local patterns within skin lesion regions. This mechanism enhances feature representation and interpretability concurrently, contributing to the robustness of the proposed network.
\item
The DAM is combined with non-homologous BCNN and CCE. The DAM guides the network to extract useful local pattern features in a weakly supervised manner, providing strong regularization to prevent overfitting when training on limited data.
\item
Thorough experiments are conducted on the HAM10000 and ISIC2019 datasets without additional training data. Results showcase the proposed network’s state-of-the-art performance in skin disease classification, demonstrating robustness and interpretability. These characteristics enhance the potential clinical application of the proposed network.
\end{itemize}

\section{Literature Review}
\label{sec:headings}
\paragraph{} Existing literature on automated skin disease classification primarily focuses on exploring various CNN architectures, feature extraction techniques, and attention mechanisms. Several studies have proposed innovative CNN-based models designed specifically for skin disease diagnosis, utilizing transfer learning, data augmentation, and ensemble methods to improve classification performance. For instance, Josphineleela et al. \cite{Josphineleela2023} proposed a multi-stage faster RCNN-based iSPLInception network to effectively classify benign and malignant tumors with addressing overfitting issues. The method utilizes faster RCNN for image classification, augmented with the iSPLInception model to reduce computation time and network complexity. Inception-ResNet is applied in iSPLInception for multi-stage classification. The prairie dog optimization algorithm is evaluated to detect the candidate box. ISIC 2019 and HAM10000 datasets are used to conduct the experimental results. Compared the accuracy, precision, recall, and F1-score with existing methods like CNN, hybrid DL, Inception v3, and VGG19. The proposed method achieves 95.82\% accuracy, 96.85\% precision, 96.52\% recall, and 0.95 F1 score, demonstrating effective prediction and classification. A bilinear CNN method is presented to classify seven skin lesion classes with high accuracy and low computational cost \cite{Camilo2021}. The method includes data augmentation to address data imbalance, transfer learning, and fine-tuning to enhance classification performance while reducing computational load. Using the HAM10000 dataset, simulations demonstrated that combining ResNet50 and VGG16 architectures improved accuracy by 2.7\% over the state-of-the-art methods. The proposed method achieved the best mean accuracy of 93.21\%, precision of 92.92\%, recall of 93.00\%, F1-score of 93.21\%, and AUC of 98.10\% by 239 minutes of training.
\paragraph{} Pundir et al. \cite{Pundir2023} introduced a skin cancer detection technique using Bilinear-CNN-SA to detect malignant moles early which provides significant patient benefits. Preprocessing steps include noise removal, grayscale conversion, and image enhancement. Otsu segmentation is applied after data cleanup. Feature extraction is done using the ABCD Rule, followed by classification using the Bilinear CNN-SA model. Comparative analysis shows superior performance compared to CNNs and Support Vector Machines (SVM) with the proposed method.
In \cite{Wei2024}, proposed a method based on Multi-Modal Bilinear Fusion (MBF) with a Hybrid Attention (HA) mechanism for skin lesion classification. MBF-HA employs a common representation learning framework to capture correlated features between modalities. Additionally, it utilizes a HA-based reconstruction module to enhance feature extraction and localize lesion regions. Experiments on the 7-point Checklist database show MBF-HA achieves an average accuracy of 76.3\% in multi-classification tasks and 76.0\% in diagnostic tasks. Nigat et al. \cite{Nigat2023} employed CNN to classify four common fungal skin diseases: tinea pedis, tinea capitis, tinea corporis, and tinea unguium. They used a dataset comprising 407 fungal skin lesions from Gerbi Medium Clinic, Jimma, Ethiopia. Image preprocessing included normalization, RGB to grayscale conversion and intensity balancing. Images were resized to $120 \times 120$, $150 \times 150$, and $224 \times 224$, followed by augmentation. Their model achieved a 93.3\% accuracy in disease classification. The proposed model achieved better results than MobileNetV2 and ResNet 50.
\paragraph{} A CNN-based fruit classification system \cite{Prakash2023} is introduced which employs bilinear pooling with heterogeneous streams. This framed fruit classification as a fine-grained visual classification and developed a heterogeneous bilinear network, then compared it with conventional approaches. The CNN network was initialized with ImageNet weights and pre-trained networks as components in the BCNN. These CNNs are used as feature extractors and combined with bilinear pooling functions. The BCNN classifier is trained and tested with Fruits-360, ImageNet, and VegFru datasets. The performance of the proposed model is validated using various metrics by comparing some existing methods, the proposed method achieved a classification accuracy of 99.69\% and an F1 score of 99.68\%, outperforming other CNN models. A method \cite{Liu2022} based on model fusion with bilinear pooling by combining EfficientNet and DenseNet is proposed. Image segmentation and feature extraction are performed using these two models separately. Bilinear features are obtained by calculating the outer product and average confluence of different spatial locations. The monkeypox skin lesion dataset is used for the binary classification of monkeypox and non-monkeypox. The model achieved the best accuracy of 94.57\%, with a recall of 94.00\% and an F1-score of 94.50\%. Detection and classification can assist in identifying monkeypox patients, which reduces the pressure of early diagnosis and prevents the monkeypox virus from spreading too fast.
\paragraph{} The authors \cite{Lin2015} introduced a BCNN-based architecture for fine-grained recognition, which represents images as pooled outer products of features from two CNNs. This captures localized feature interactions in a translationally invariant manner. BCNNs are orderless texture representations that can be trained in an end-to-end manner. The most accurate models achieved per-image accuracies of 84.1\%, 79.4\%, 84.5\%, and 91.3\% on the Caltech-UCSD Birds, NABirds, FGVC Aircraft, and Stanford Cars datasets. Additionally, the method systematically analyzed these networks and showed that: a) the size of the bilinear features can be significantly reduced without loss of accuracy, b) these are also effective for other tasks such as texture and scene recognition, and c) it can be trained from scratch on ImageNet with consistent improvements over the baseline. Visualizations of these models on various datasets are presented using top activation and gradient-based inversion techniques. Inthiyaz et al. \cite{Inthiyaz2023} proposed an automated image-based method utilizing machine learning to diagnose and categorize skin disease problems. Computational techniques are employed to analyze, process, and organize picture data, considering various image characteristics. Skin images undergo noise removal and quality enhancement processes. Features are extracted using CNN, images are classified using the Softmax classifier algorithm, and a diagnostic report is generated. This system could serve as a dependable real-time educational tool for dermatology researchers.
\paragraph{} In \cite{Karambele2024}, a novel approach, eDermaCare is introduced, which utilizes ResNet-50 for skin disease classification. In contrast, other CNN architectures produced unsatisfactory results, ResNet-50 captures complex skin features well. The focus was on enhancing classification through DL capabilities. The study explores training and retention challenges, highlighting block balances to prevent performance degradation. Ultimately, the model achieves 76.07\% classification accuracy. Sazzadul et al. \cite{Sazzadul2023} aimed to diagnose Eczema and Psoriasis using deep CNN architectures. Five state-of-the-art CNN architectures were employed, and their performance was assessed using 10-fold cross-validation. Inception ResNet v2 with the Adam optimizer achieved a maximum validation accuracy of 97.1\%. The study presents two practical application approaches: a) a smartphone-oriented approach, integrating CNN models with a mobile application; and b) a web server-oriented approach, integrating the CNN model with a web server for real-time skin disease classification.
Gautam et al. \cite{Gautam2023} attempted various techniques to detect skin diseases early but obtained unsatisfactory results. Their study aims to classify, detect, and provide accurate information about skin diseases. They propose a high-performance CNN to achieve this goal. The methodology involves preprocessing skin disease images, extracting important features, and analyzing them using a deep CNN. Their approach achieves up to 98\% accuracy in detecting six different diseases, being simple, fast, and accurate.

\section{Materials and Methods}
\paragraph{} This paper introduces a non-homologous BCNN for skin disease classification. The model comprises three key components: a feature extraction module, a DAM module, and a compact bilinear pooling module. The feature extraction module is built upon ResNet-50 \cite{Pan2010} and Xception \cite{Chollet2017}. Image features extracted by these networks form a comprehensive data feature map, further enhanced by the DAM. The resulting data feature maps from both CNNs are then directed to the FC layer for classification following Compact Bilinear Pooling (CBP). The network notably enhances feature extraction capability and classification performance by combining two CNNs with distinct architectures and integrating a DAM module.

\subsection{Bilinear-CNN Architecture}
\paragraph{}A BCNN architecture consists of two equal or different CNNs (streams A, and B) in a parallel manner. An input feeds both CNNs and a training step computes the weights for each neuron and the tensor output of two models for the features at each image location and then pooling to obtain the image descriptor. The outer product and sum pooling for two CNN tensor outputs combine feature vector at each location, i.e., for any location l of $i^{th}$ image is denoted as:

\begin{equation}
\phi\left( c,\ f_{A},f_{B} \right) = \sum_{i = 1}^{p,q}{U_{c,p,q,f_{A}}{\ V}_{c,p,q,f_{B}}}
\label{eq:1}
\end{equation}

where \(U \in \mathbb{R}^{c \times p \times q \times f_{A}}\) and \(V \in \mathbb{R}^{c \times p \times q \times f_{B}}\) are the result matrix from \(f_{A}\) and \(f_{B}\), and c denotes the batch size, \emph{p}, and \emph{q} the spatial dimensions of \(f_{A}\) and \(f_{B}\). The resulting matrix \(\phi \in U^{(c \times f_{A} \times f_{B})}\) captures pairwise correlations between feature channels. The matrix offers richer representations compared to linear models. After reshaping the output matrix \(\phi\) to \(c \times f_{A} \times f_{B}\), it is normalized by signed square root normalization as given below:

\begin{equation}
R = \text{sign}(\phi) \sqrt{|\phi|}
\label{eq:2}
\end{equation}

where R is the signed square root matrix which is normalized by the \emph{L2} norm to obtain the final value that passes to the FC layer as:

\begin{equation}
S = \frac{R}{\left\| R \right\|^{2}}
\label{eq:3}
\end{equation}

\paragraph{} Finally, the output of the model passes through an FC layer with softmax activation to derive predictions. The lack of large-scale labeled datasets can make it challenging to train a DCNN from scratch, especially in the medical image classification domain. Transfer learning is an alternate method that uses a pre-trained model to start a new task. This method enables quick development and improved modeling performance for the next task.
\paragraph{} The BCNN adopted in this study involves integrating a ResNet-50 \cite{Pan2010} in one arm and the Xception network \cite{Chollet2017} in the other. FC layers are removed from each model before their combination into the BCNN. Following an outer product operation, a single FC layer is adjusted for the classification, incorporating seven neurons instead of the original 1000.
\paragraph{} ResNet-50 and Xception models, pre-trained on ImageNet \cite{He2015} and available in the Keras library, are employed for this study. A seven-neuron layer replaces the FC layer with 1000 neurons, and to retrain the weights for the target task without random initialization, the average pooling layer is removed. ResNet-50, a DCNN, has demonstrated high performance in computer vision tasks, clinching the ImageNet challenge in 2015. Noteworthy for learning residual representation functions instead of direct signal representation, ResNet employs skip connections to combat vanishing gradient issues, promoting model convergence. Specifically chosen for its superior performance in this context, ResNet-50 initiates with a 7 × 7 kernel size for the initial convolution, followed by Max-pooling with a 3 × 3 kernel size. The model consists of four stages, each with 2, 4, 6, and 3 residual blocks. Each residual block comprises three convolutional layers with kernel sizes of 1 × 1, 3 × 3, and 1 × 1. The 1 × 1 convolution layer serves for dimension reduction and restoration, while the 3 × 3 layer acts as a bottleneck with smaller input/output dimensions. Concluding the architecture is an average pooling layer followed by an FC layer with 1000 neurons. In total, ResNet-50 boasts over 23 million trainable parameters.
\paragraph{} Conversely, Xception is a deep CNN boasting 36 convolutional layers, surpassing ResNet in the ImageNet challenge. With a total of up to 22 million parameters, the Xception model consists of 14 modules arranged into three stages: the entry flow, the middle flow, and the exit flow. The entry flow starts with four modules. The first module has an input layer with dimensions of 299 × 299, and then there are two convolutional layers with kernel sizes of 3 × 3. Modules that come after that have two separable convolutional layers with a 3 × 3 kernel size, a 3 × 3 max pooling layer, and a linear residual connection coating every module. A middle flow iterates eight times over a module with three separable convolutional layers, each time using a 3 × 3 kernel size and incorporating a residual connection around the module. Two separable convolutional layer modules with a 3 × 3 kernel size, a 3 × 3 max pooling layer, and a residual connection are the first components of the exit flow. Finally, a Global Average Pooling (GAP) layer is used before two separable convolutional layers. Fine-tuning is applied to enhance performance by freezing 70\% of layers for both the ResNet-50 and Xception models. The weights of the remaining layers are adjusted, allowing for meaningful improvements by iteratively adapting pre-trained features to the new data.

\subsection{Compact Bilinear Pooling and Non-Homologous BCNN}

\paragraph{} Traditional CNNs are unable to extract distinct features from the input skin disease images due to their fine-grained nature. For this reason, the bilinear pooling \cite{Lin2015} described in this paper works well for classifying skin disease images. The BCNN treats features from two different information streams as unique and combines high-order statistical data. The model effectively captures connections between the two streams by performing outer product and pooling operations on these features, yielding global features. This approach leverages image translation invariance to model local dual features dynamically. The discriminative features obtained through BP are then used for image classification for better results. Two CNNs with differing structures are ultimately employed to extract various image features.
\paragraph{} While bilinear pooling performs well in image classification, the bilinear features generated are often high-dimensional, ranging from hundreds of thousands to millions. Such ineffective representations significantly increase memory usage, rendering them unsuitable for subsequent analysis. Compact BCNN is employed to conduct core analysis on bilinear features to reduce the dimension. Compact BCNN retains the discriminative capacity of Bilinear Pooling (BP) but with reduced dimensions in the thousands. It enables the backpropagation of misclassified samples, facilitating end-to-end optimization of image classification. A compact bilinear representation is derived through core analysis, relying on low-dimensional feature maps of the kernel function. The bilinear pooling model forms a global image descriptor through a sequence of the following calculations:

\begin{equation}
blinear\left( i, l, f_{A}, f_{B} \right) = f_{A}(i, l)^{T}  f_{B}(i, l)
\label{eq:4}
\end{equation}

\paragraph{} It is essential to appropriate feature representation for a specific model, which requires feature vectors \(f_{A}\), and \(f_{B}\) to be equal feature dimensions. The pooling function generates the global feature representation \(\phi(l)\) for an image by accumulating the bilinear feature combinations at every location of an image. In our method, we used pooling sum as shown below:

\begin{equation}
\phi(l) = \sum_{l \in L}^{}{bilinear \left( i,l,f_{A},f_{B} \right)}
\label{eq:5}
\end{equation}

\paragraph{} The global feature representation $\phi(l)$ for an image is order-less due to ignoring the feature locations by pooling; hence, for every spatial location $l$ in a location space $L$ of image $i$, we compute the bilinear feature vector. If $f_{A}$ and $f_{B}$ extract corresponding local feature vectors of size $Y \times N$ and $Y \times M$, then their bilinear combination $\phi(l)$ will have size $N \times M$. The similarity function $S$ uses the general-purpose bilinear vector representation. Instinctively, the outer product indicates the pairwise interactions of the feature extractors $f_{A}$ and $f_{B}$ on each other. To appropriate vector representation of a particular model, $f_{A}$ and $f_{B}$ need to be equal feature dimension $Y$ as $f: l \times L \rightarrow R^{Y \times D}$ of size $Y \times D$, where $l \in L$ represents the location of any image $i \in l$. The pooling function generates the global feature representation $\phi(l)$ of image $i$ by accumulating the bilinear feature combinations at every location of an image as formulated by Equation \ref{eq:5}. $\phi(l)$ symbolizes a vector projection with low dimensions; $f_{A}$ and $f_{B}$ represent feature vectors. If we represent the second-order polynomial kernel function using $\mathit{k}\left(f_{A}, f_{B}\right)$, then, to make the value of $d$ considerably lower than $c^{2}$, we can find a low-dimensional projection function $\phi\left( f_{A} \right) \in R^d$ that fulfills Equation \ref{eq:6}:

\begin{equation}
\phi\left( f_{A} \right),\ \phi(f_{B}) \approx \mathit{k} \left( f_{A},f_{B} \right)
\label{eq:6}
\end{equation}

where $\mathit{k}\left(f_{A}, f_{B}\right)$ denotes the low-dimensional vector, and $\phi\left( f_{A} \right) \in R^{d}$ denotes a projection function. The $c^2$-dimensional inner product of the original is approximately equal to the inner product reduced to the $d$-dimension. The representation that follows can be obtained by:

\begin{equation}
\begin{split}
\left( \mathfrak{b}(A), \mathfrak{b}(B) \right) = \sum_{l \in L}^{}{{{{(f}_{A}(i,l),\ f}_{A}(i,l)}^{T})}\sum_{l \in L}^{}{{{(f_{B}(i,l),\ f}_{B}(i,l)}^{T})} \\
 = \sum_{i \in l}^{}{\sum_{l \in L}^{}\left( \phi\left( f_{A} \right),\ \phi\left( f_{B} \right) \right)} = (\upsilon(A),\ \upsilon(B))
\label{eq:7}
\end{split}
\end{equation}

where $A$ and $B$ are the set of local descriptors, $l$, and $L$ is the set of spatial positions, $A_{i}$ and $B_{l}$ represent local descriptors, and $\upsilon(A)$ and $\upsilon(B)$ represent low-dimensional vectors after $\mathfrak{b}(A)$ and $\mathfrak{b}(B)$ are mapped by the mapping function; where $\upsilon(A) = \sum_{i \in l}^{}{\phi(A_{i}),}$ and $\upsilon(B) = \sum_{l \in L}^{}{\phi(B_{l})}$ are the compact bilinear features, $i$ and $l$ are the number of channels of the vector $\mathfrak{b}(A) = \sum_{i \in l}^{}{{f_{A}(i,l),\ f}_{A}(i,l)}^{T}$, and $\mathfrak{b}(B) = \sum_{l \in L}^{}{{f_{B}(i,l),\ f}_{B}(i,l)}^{T}$. Any polynomial kernel can provide a low-dimensional approximation function to reduce dimensionality.

\subsection{Dual Attention Mechanism}

\begin{figure}[!htbp]
\centering
{\includegraphics[scale=0.60]{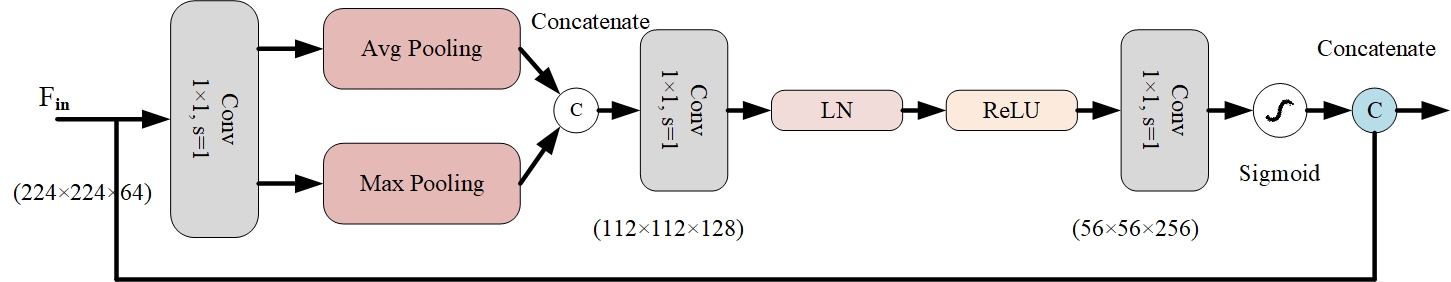}}
\caption{Improved CAM mechanism.}
\label{Figure1}
\end{figure}

\paragraph{} In addition to embedding an attention mechanism with ResNet-50 and Xception for the weight allocation of the input features, the Convolutional Block Attention Module (CBAM) \cite{Woo2016} is further improved in this paper. It proposes a new attention mechanism to enhance the accuracy of ResNet-50 with Xception in performing classification to the fullest extent. This section introduces the improved CBAM in detail.
\paragraph{} Building upon the BCNN, this paper introduces an enhancement to the CBAM by incorporating bi-linearity principles, resulting in a bilinear CBAM. Moreover, the feature information from channels and spatial dimensions is integrated to reinforce the CAM within the CBAM. To further refine the algorithm, the linearity of the BCNN is considered, retaining the original maximum pooling and average pooling in parallel to preserve the initial map's feature information comprehensively. The output result concatenation is subsequently executed based on the channels. The FC layer in the original structure transforms into a $1 \times 1$ convolutional layer with layer normalization applied for batch data processing. Given that both ResNet-50 and Xception utilize the ReLU activation function consistently, the CAM’s ReLU function is incorporated with random regularity. The structural representation of this configuration is depicted in Figure \ref{Figure1}.

\paragraph{} The input data processing by the above-mentioned network is shown in Equation \ref{eq:8}:

\begin{equation}
\begin{split}
F_{A1} = \ \mu (Conv(ReLU(LN(Conv(Concat(AP(F),MP(F))))))) \\
 = \mu (W1(ReLU(LN(W_{0}(Concat(F_{avg}^{c},\ F_{\max}^{c}))))))
\label{eq:8}
\end{split}
\end{equation}

where $\mu$ and ReLU are activation functions, $Conv$ is the convolution, $AP$ is the Average Pool, $MP$ denotes the Max Pool, and $F_{A1}$ is the feature map produced by the CAM. $F \in R^{(B, 2C/r, W, H)}$ follows the application of the ReLU activation function, while $F \in R^{(B, C, W, H)}$ follows the application of the sigmoid activation function.

\begin{figure}[!htbp]
\centering
{\includegraphics[scale=0.56]{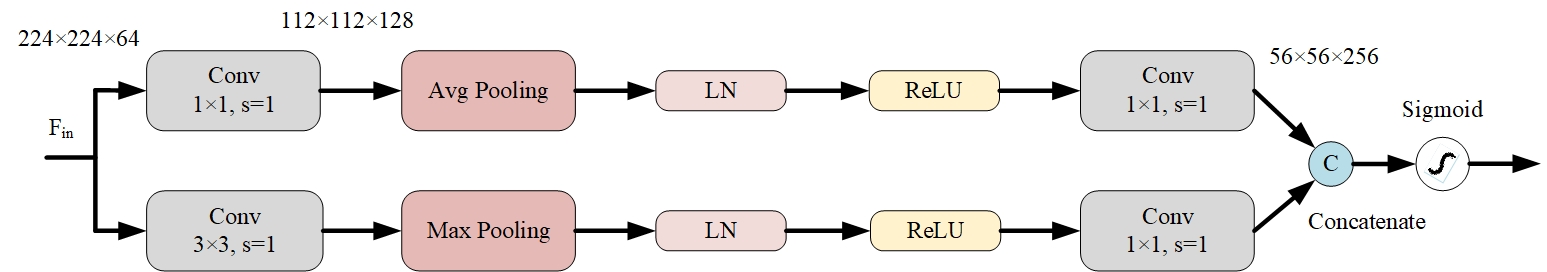}}
\caption{Improved SAM mechanism.}
\label{Figure2}
\end{figure}

\paragraph{} Moving forward, the Spatial Attention Module (SAM) is infused with linearity. He et al. \cite{He2020} introduced a spatial attention method utilizing $1 \times 1$ and $3 \times 3$ convolutions to extract diverse and abundant feature information. In this paper, we build upon this method for enhancement. Initially, $1 \times 1$ and $3 \times 3$ convolution operations are conducted on the input feature map to obtain spatial features of varying scales. Subsequently, layer normalization and ReLU mappings are applied to the two features. The dimensions of the features are then reduced to single channels using $1 \times 1$ convolution. Multiplying these two single-channel features produces multi-scale spatial attention features. These features are further multiplied by the feature map with channel attention weight, resulting in a feature map encompassing both channel weight and spatial features. Batch processing is implemented for data using layer normalization for two primary reasons. First, this approach aligns with the data processing methodology of ResNet-50 with Xception, building upon their research results. Second, it follows the consistent use of batch normalization in CNNs, with the added benefit of independence from the hardware constraints associated with batch size.
\paragraph{} In selecting convolution kernel sizes, $1 \times 1$ and $3 \times 3$ convolutions are employed. Beyond capturing spatial features at different scales, using $1 \times 1$ convolution reduces the overall network’s complexity and computational load. Furthermore, GPUs often optimize the analysis of $3 \times 3$ convolutions, contributing to improved efficiency. Optimizations and modifications are executed based on the considerations outlined above. The improved structure chart of the SAM is depicted in Figure \ref{Figure2}.

\paragraph{} The input data is processed by the network described above as Equations \ref{eq:9}-\ref{eq:11}:

\begin{equation} {F}_{A2} = Sigmoid\left( F_{1\ } \oplus \ F_{2} \right)
\label{eq:9}\end{equation}

\begin{equation}{F}_{1\ } = {Conv}_{\_(1 \times 1)}(ReLU(LN({Conv}_{\_(1 \times 1)}(F))))\label{eq:10}\end{equation}

\begin{equation}F_{2\ } = {Conv}_{\_(1 \times 1)}(ReLU(LN({Conv}_{\_(3 \times 3)}(F))))\label{eq:11}\end{equation}

where $F_{A2}$ is the feature map generated by the SAM; $Conv_{(1 \times 1)}$ is a convolution with a $1 \times 1$ convolution kernel; $Conv_{(3 \times 3)}$ is a convolution with a $3 \times 3$ convolution kernel; $F \in R^{(B,C/r,W,H)}$ follows the ReLU, $F \in R^{(B,1,W,H)}$ follows the sigmoid, and $\oplus$ denotes the concatenation operation.

\begin{figure}[!htbp]
\centering
{\includegraphics[scale=0.41]{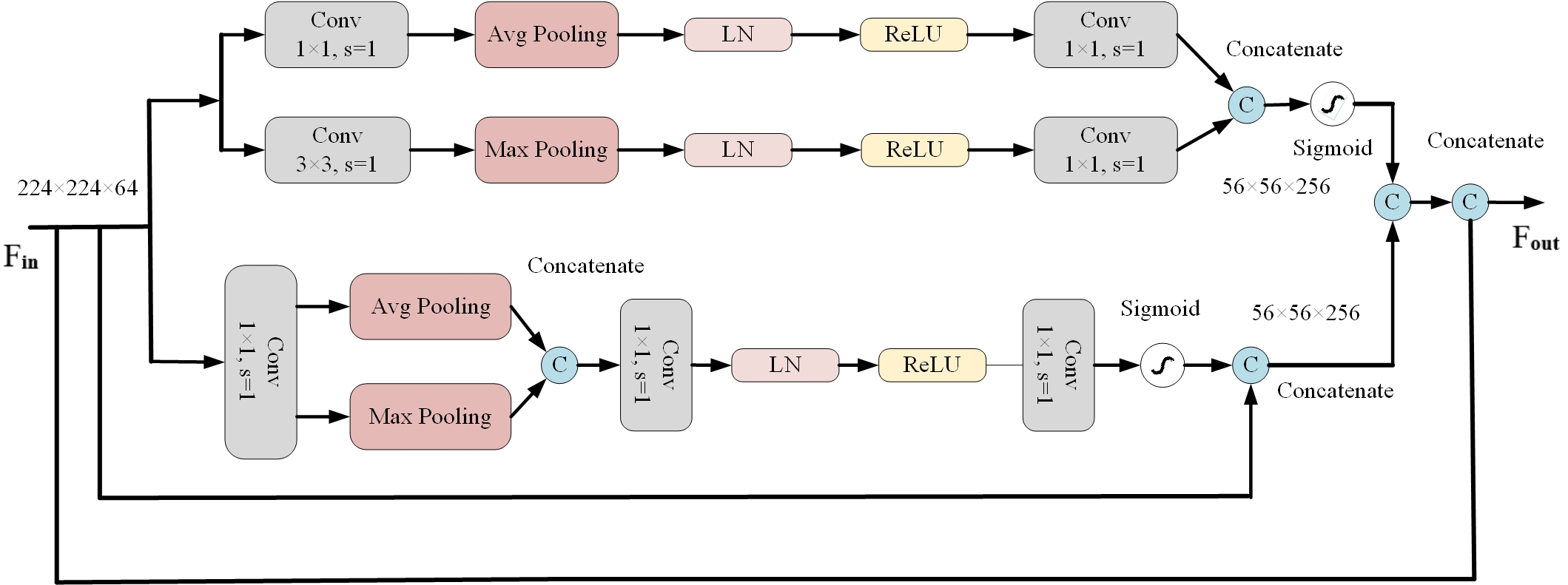}}
\caption{Structure of dual attention mechanism.}
\label{Figure3}
\end{figure}

\paragraph{} The conventional attention mechanism exclusively assigns weights between channels without considering the spatial dimension, so the DAM is constructed with two branches. The first branch provides channel weights, while the second offers spatial weights (aimed at giving channel weights). The initial convolutional DAM employs a serial processing method. In improving this mechanism, particularly for classification tasks, the network is more precise by extracting feature attention. Consequently, the serial processing approach replaces parallel processing, enabling simultaneous channel and spatial attention training. This adjustment ensures the retention of sufficient semantic information while acquiring multi-scale features for the target. Figure \ref{Figure3} shows the specific structure of the attention network The DAM, as mentioned above, can be summarized by:

\begin{equation}F_{o}\  = \ (F_{A1}\ \oplus \ F_{in})\ \oplus \ F_{A2}\ \oplus \ F_{in}\label{eq:12}\end{equation}

where $F_{in}$ represents the input feature map of BCDAM, $F_{o}$ represents the feature map of the output BCDAM, and $F_{A1}$ and $F_{A2}$ represent the feature map generated by the CAM and SAM, and $\oplus$ denotes the concatenation operation.

\subsection{DACB-Net with CCE for Skin Disease Classification}

\paragraph{} A method of two embedding attention mechanisms with ResNet-50 and Xception is proposed in this paper. A detailed implementation is conducted on the specific attention framework for this position to thoroughly investigate the embedding position of the attention mechanism within the entire network. This section introduces a novel multi-perspective attention framework. The shortcut attention branch incorporates the DAM proposed in this paper and the feature map size passing through each module.
\begin{figure}[!htbp]
\centering
{\includegraphics[scale=0.45]{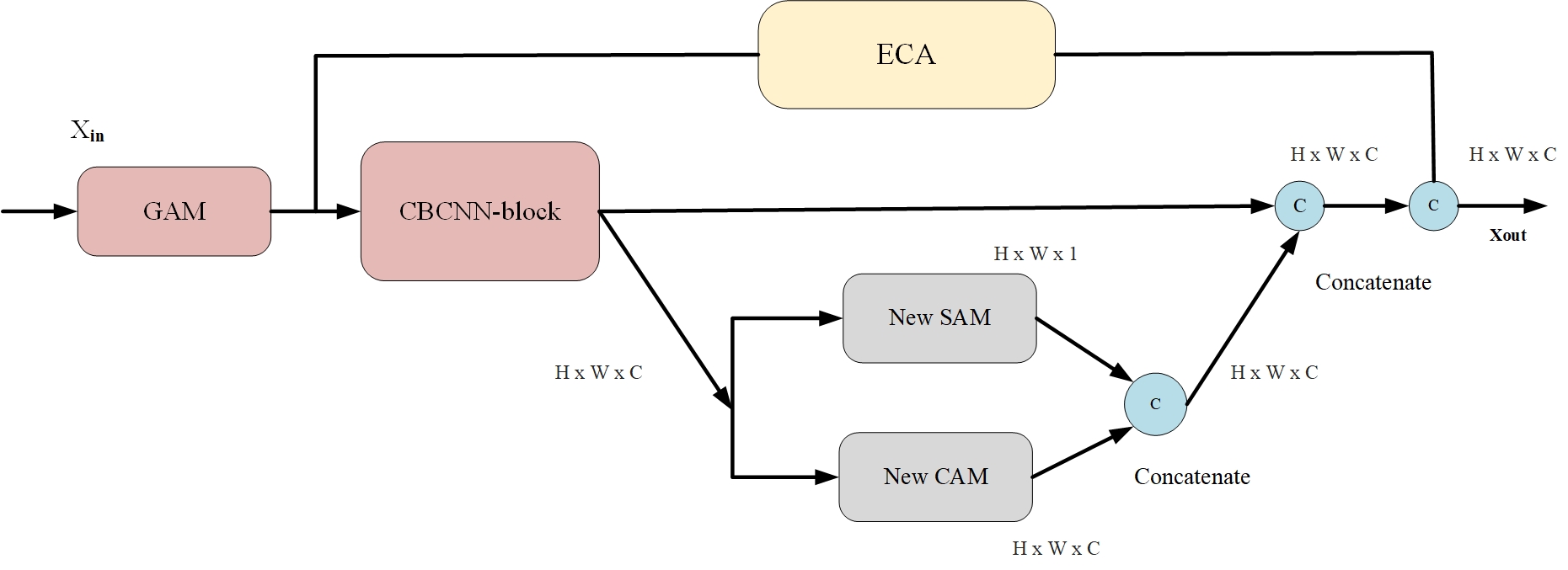}}
\caption{Bilinear attention mechanism model.}
\label{Figure4}
\end{figure}

\paragraph{} The Efficient Channel Attention (ECA) combined with the attention architecture shown in Figure \ref{Figure4} considers each channel, the spatial dimension of the feature map, and the information interaction between channels \textbackslash cite\{Wang2020\}. Due to the dimension changes in convolution, ECA covers up for the lost interaction information between channels. Using a feature map that does not cycle network blocks, this method balances relationships along various feature map directions. It retains a significant degree of channel interaction information from the original feature map, enabling the attention mechanism to address deficiencies during training and achieve comprehensive and multi-perspective attention parameter training.
\begin{figure}[!htbp]
\centering
{\includegraphics[scale=0.27]{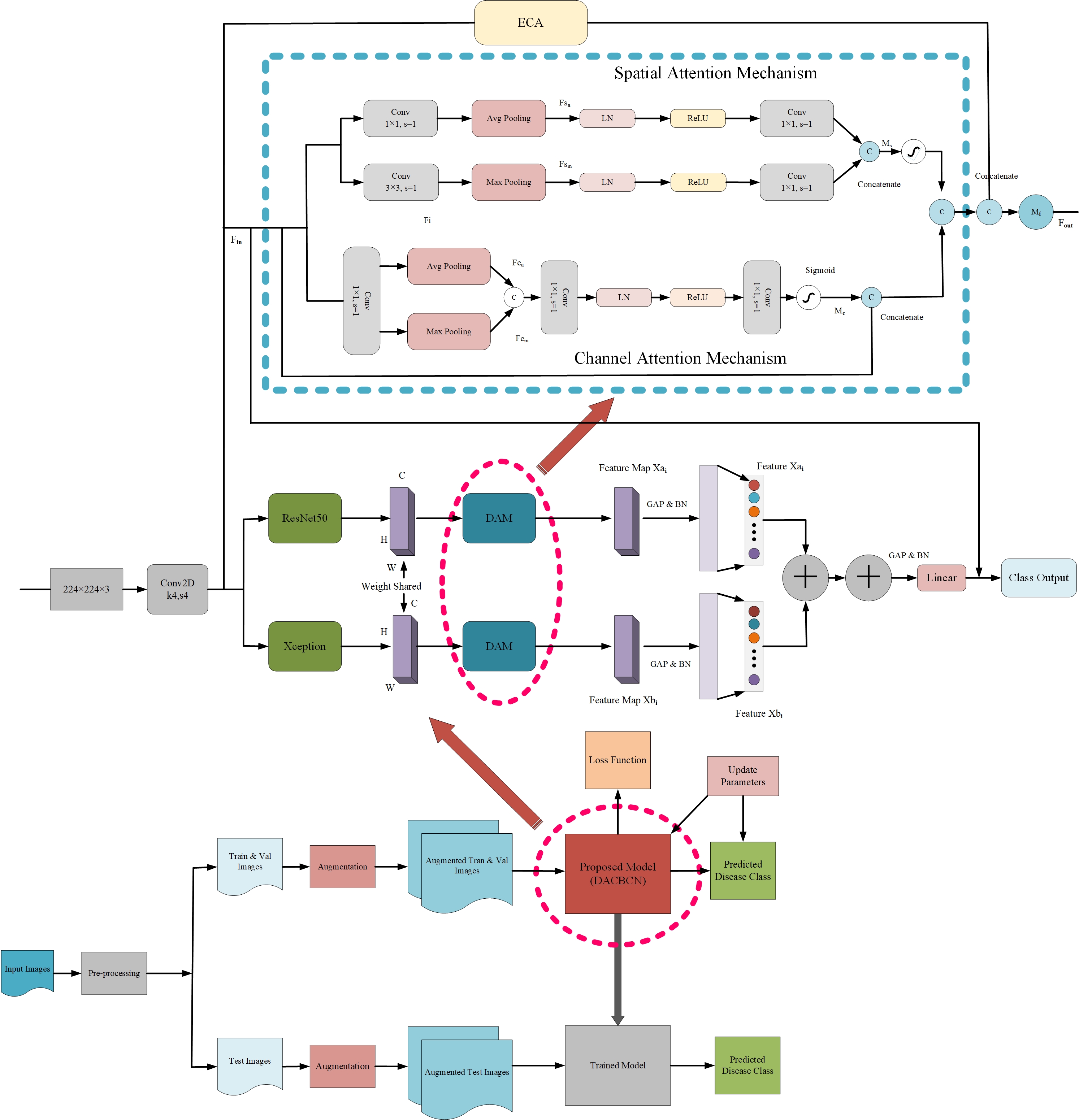}}
\caption{Proposed DACB-Net architecture.}
\label{Figure5}
\end{figure}

\paragraph{} Furthermore, when elevating the original ResNet-50 with Xception to construct a deeper architecture, this architecture aggregates the feature map with the attention output before the input block, which takes shape in channel relationships. This aids the network in extracting and focusing on sample features. Additionally, the ReLU activation function is omitted after adding the original feature map and attention feature map, contributing to an enhancement in the final classification accuracy of the network. The ReLU activation function is also removed in the subsequent comparison experiment network. Figure \ref{Figure5} indicates how the DACB-Net proposed in this paper is embedded into the ResNet-50 with the Xception network through SBA.

\subsection{Unbalanced Sample Distribution Problem}

\paragraph{} In clinical medicine, a minuscule proportion of patients exhibit serious illnesses compared to the overall number of individuals examined. This leads to a data set with a pronounced imbalance, presenting a significant distribution challenge. The class with more samples dominates the training process when using imbalanced data for network learning \cite{Inthiyaz2023}. Consequently, the trained classification models often perform well in classes with a large sample size but poorly in classes with a smaller sample size. To address the degradation of network performance due to data imbalance in datasets, the Complement Cross Entropy (CCE) is used. CCE tackles the problem by leveraging the training of non-labelled category samples. It eliminates the need to increase the sample count or the size of minority losses. Instead, this method utilizes information from incorrect classes to robustly train a classification model under unbalanced class distribution.
Consequently, this strategy offers improved learning opportunities for categories with a limited sample size. It achieves this by encouraging the correct category (including a few categories) to have a higher Softmax score than all other incorrect categories. The CCE loss function is calculated from the average value of each batch's sample entropy of the wrong class. The CCE is defined as Equation \ref{eq:13}:

\begin{equation}
L(y,\widehat{y}) = - \frac{1}{N}\sum_{i = 1}^{N}\mspace{2mu}\sum_{j = 1,j \neq g}^{K}\mspace{2mu}\frac{{\widehat{s}}^{(i)\lbrack j\rbrack}}{1 - {\widehat{s}}^{(i)\lbrack g\rbrack}}\log\frac{{\widehat{s}}^{(i)\lbrack j\rbrack}}{1 - {\widehat{s}}^{(i)\lbrack g\rbrack}}
\label{eq:13}
\end{equation}

where $g$ is the sequence number of the ground truth category, $\widehat{s}(i){[}j{]}$ is the prediction of the $i^{th}$ sample in class $g$, $N$ is the number of samples in the mini-batch, $P$ is the total number of categories, $s$, and $\widehat{s}$ are the feature map and its estimated probability vectors of the given sample. This entropy aims to predict the probability that the ground truth class will have greater significance than the wrong classes. This is achieved by distributing the Softmax scores among the incorrect classes. In other words, there is higher confidence in guessing the true class when there is a more neutral prediction probability distribution for the false class. Since the complement entropy achieves its maximum when the probability distribution is uniform, the optimizer aims to maximize it. The complementary entropy and cross-entropy scales align with the balanced complement entropy is defined as:

\begin{equation}
\widehat{L}\left( s,\widehat{s} \right) = \frac{1}{P - 1}L\left( s,\widehat{s} \right)
\label{eq:14}
\end{equation}

where the balancing coefficient is $1 / (P-1)$, the adjustment factor $g$ is added to balance the cross entropy and complement entropy. Equation \ref{eq:15} shows the final loss function:

\begin{equation}
C =  \widehat{H}\left( s,\widehat{s} \right) + \frac{\beta}{P-1}L(s,\widehat{s}).
\label{eq:15}
\end{equation}

\paragraph{} The advantage of loss over focal loss is that its internal connection may be balanced without the need to establish weights. The loss function can solve the problem of sample imbalance by examining the relationship within the sample.

\section{Dataset and Evaluation}
\subsection{Dataset Acquisition and Description}
\paragraph{} The proposed method is evaluated on two publicly available benchmark datasets: HAM10000 \cite{Tschandl2019} and ISIC2019 \cite{Kassem2020}. The details of the datasets are described in the following.

\begin{figure}[!htbp]
\centering
{\includegraphics[scale=0.46]{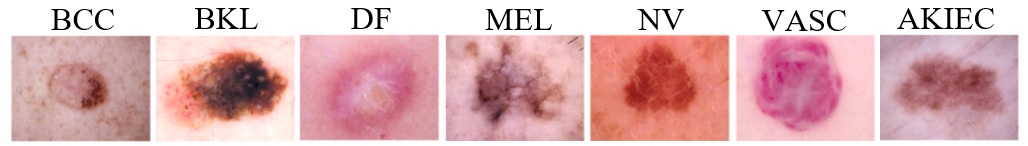}}
\caption{Seven skin lesion classes in the HAM10000 dataset.}
\label{Figure6}
\end{figure}

\paragraph{HAM10000:} It consists of a total of $10015$ dermoscopic images. Seven types of skin lesions are included in this dataset as shown in Figure \ref{Figure6}: $BCC$, $BKL$, $DF$, $MEL$, $NV$, $VASC$, and $AKIEC$. However, all classes are not balanced; therefore, we performed data augmentation for the ﬁnal classiﬁcation. This dataset consists of $514$ in $BCC$, $1099$ in $BKL$, $115$ in $DF$, $1113$ in $MEL$, $327$, $6705$ in $NV$, $142$ in $VASC$, and $327$ images in $AKIEC$ class. The number of images shows that the dataset is unbalanced. Therefore, we augmented the input data before training. In this work, we use 70\% of the images for training, the rest 15\% for validation, and 15\% for testing. Select all the images in a randomized process.

\paragraph{ISIC2019:} There are $25331$ labelled images of $8$ categories in the ISIC2019, which split into $867$ Actinic keratosis (AK), $3323$ Basal cell carcinoma (BCC), $2624$ Benign keratosis (BKL), $239$ Dermatofibroma (DF), $4522$ Melanoma (MEL), $12875$ Melanocytic nevus (NV), $628$ Squamous cell carcinoma (SCC), and $253$ Vascular lesion (VASC). We divide each class of the ISIC2019 into three parts with identical data distributions, in terms of 80\% for training data, 10\% for validation data, and 10\% for test data. So the training, validation, and testing parts own $20269$, $2531$, and $2531$ images, respectively.

\begin{table*}[!htbp]
\center
\setlength{\tabcolsep}{5pt}
\caption{Dataset information of HAM10000 and ISIC2019.}
\label{Table 1}
\begin{tabular}{|*{4}{c|}}
\hline
\rule{0pt}{1.0\normalbaselineskip}
\bf{Dataset} & \bf{Class} & \bf{Before} \bf{(Data Augmentation)} & \bf{After} \bf{(Data Augmentation)} \\[5pt]
\hline
\rule{0pt}{1.0\normalbaselineskip}
\multirow{7}{*}{HAM10000} & Basal Cell Carcinoma (BCC) & 514 & 1750 \\[5pt]
& Benign Keratosis (BKL) & 1099 & 1750 \\[5pt]
& Dermatofibroma (DF) & 115 & 1750 \\[5pt]
& Melanoma (MEL) & 1113 & 1750 \\[5pt]
& Melanocytic Nevus (NV) & 6705 & 1750 \\[5pt]
& Vascular Lesion (VASC) & 142 & 1750 \\[5pt]
& Actinic Keratosis (AKIEC) & 327 & 1750 \\[5pt]
\hline
& \bf Total & 10015 & 12250  \\[5pt]
\hline
\multirow{8}{*}{ISIC2019} & Actinic Keratosis (AKIEC) & 867 & 1750 \\[5pt]
& Basal Cell Carcinoma (BCC) & 3323 & 1750 \\[5pt]
& Benign Keratosis (BKL) & 2624 & 1750 \\[5pt]
& Dermatofibroma (DF) & 239 & 1750 \\[5pt]
& Melanoma (MEL) & 4522 & 1750 \\[5pt]
& Melanocytic Nevus (NV) & 12875 & 1750 \\[5pt]
& Squamous Cell Carcinoma (SCC) & 253 & 1750 \\[5pt]
& Vascular Lesion (VASC) & 628 & 1750 \\[5pt]
\hline
& \bf Total & 25331 & 14000 \\[5pt]
\hline
\end{tabular}
\end{table*}

\subsection{Dataset Pre-Processing and Augmentations}

\begin{figure}[!htbp]
\centering
{\includegraphics[scale=0.30]{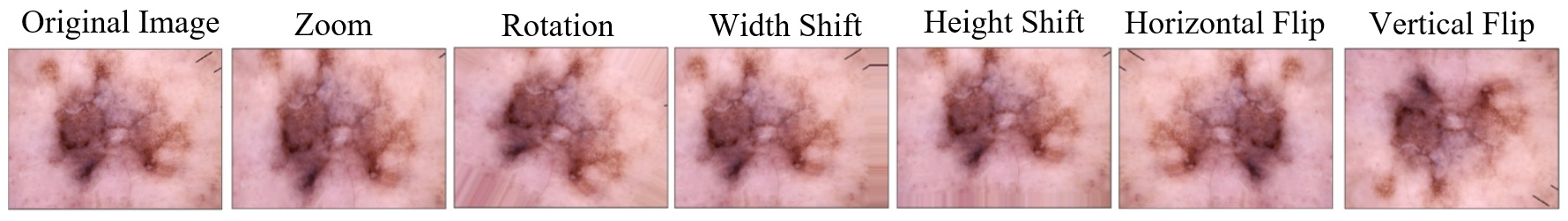}}
\caption{Data augmentation results with (a) Zoom, (b) Rotation, (c) Width Shift, (d) Height Shift, (e) Horizontal Flip, (f) Vertical Flip of the HAM10000 dataset.}
\label{Figure7}
\end{figure}

\paragraph{} As previously mentioned, some classes in the dataset have low sample sizes. Therefore, data augmentation is needed to solve this problem. However, before that, the dataset was split into a train-test-split, and augmented samples were not included in the test set.
\paragraph{} For training, 80\% of the data was used, and the remaining 80\% for testing. A total of 6 augmentation techniques, which were Rotation, Zoom, Vertical Flip, Horizontal Flip, Brightness, and Shear, were performed for the training set as shown in Figure \ref{Figure7}. Table \ref{Table 1} shows the number of input samples taken from each class before and after augmentation for the HAM10000 and ISIC2019 datasets, and the total number of samples used for each splitting technique is shown in Table \ref{Table2}.

\begin{table*}[!htbp]
\center
\caption{Number of samples for all splits.}
\label{Table2}
\renewcommand{\arraystretch}{1.5}%
\begin{tabular}{|*{5}{c|}}
    \hline
\multirow{2}{*}{\bf{Splitting and Augmentation}} & \multicolumn{2}{c|}{\bf{No. of Train Samples}} & \multicolumn{2}{c|}{\bf{No. of Test Samples}} \\\cline{2-5}
                          & \bf{HAM10000} & \bf{ISIC2019} &  \bf{HAM10000} &  \bf{ISIC2019} \\ 
\hline
(Without Augmentation) &   7010          &    17732   &      3305         &    7599   \\ 
\hline
(With Augmentation)      &   8575          &     9800    &      3675         &    4200   \\
\hline
(Without Augmentation) &   8012          &     20265    &     2003       &    5066  \\
\hline
(With Augmentation)     &    9800          &      11200   &     2450        &    2800  \\
\hline
\end{tabular}
\end{table*}

\subsection{Model Training}
\paragraph{} Six models trained on two dataset splits were collected in the previous steps for the present study. The proposed model reached convergence in about 320 minutes on the NVIDIA GTX 980 Ti GPU occupying less than 4GBs of GPU memory to accelerate the computation for full implementations Each model was trained to 50 epochs to evaluate which model performed well. Learning rate and weight decay change depending on the model's architecture. We use the Adam \cite{Kingma2014} optimizer with default momentum values (0.9, 0.999) for $\beta1$ and $\beta2$. The weight decay is 0.0001. The learning rate initializes to $1 \times 10^{-3}$ for both HAM10000 and ISIC2019 datasets. The proposed model was the best-scoring model on 50 epochs. The extensive training took the previous testing accuracy of 50 epochs from 92.86 \% to 94.24 \%.

\subsection{Classification Metrics}
\paragraph{} Initially, we computed the confusion matrix to get all classes TP, TN, FP, and FN scores. All four measures are calculated as one against other categories. Six out of the seven behaved as negative class for HAM10000. Similarly, seven out of the eight are considered to be the negative class for ISIC 2019. If the score is computed for any class. TP denotes the true positive and accurate prediction for any class the model makes. FP denotes the false positive and false prediction for any class. Similarly, TN and FN represent true negative and false negative. These metrics are computed and combined to comprehensively evaluate the model's performance, such as precision, recall, F1-score, and accuracy.
\paragraph{} Evaluate the performance using early-discussed measures, such as precision, recall, F1 score, and accuracy. The total counts of true positive are divided by the overall true positive + false positive and calculated by Equation \ref{eq:16}:

\begin{equation}
\ \ \ \ \ Precison = \frac{TP}{TP + FP}.
\label{eq:16}
\end{equation}

\paragraph{} To compute recall, divide the number of true positives by the total number of true positives and false negatives, as computed in Equation \ref{eq:17}:

\begin{equation}
\ \ Recall = \frac{TP}{TP + FN}.
\label{eq:17}
\end{equation}

\paragraph{} The F1-score is the mean of precision and recall when both metrics are used is shown as:

\begin{equation}
\ F1 - score = \frac{2 \times (Precision \times Recall)}{Precision\  + \ Recall}.
\label{eq:18}
\end{equation}

\section{Experiments and Results}
\paragraph{} This section assesses the effectiveness of the skin lesion classification method. Specifically, we categorize lesions from the HAM10000 and ISIC2019 datasets, apply data augmentation techniques, and quantitatively assess performance using six standard metrics in skin disease classification. Ultimately, we compare the results we obtained against other state-of-the-art methods \cite{Srinivasu2021} \cite{Kassem2020} \cite{Raju2022} \cite{Bharat2022} \cite{Kumar2022} \cite{Brinker2022} \cite{Khan2018} \cite{Indraswari2022} \cite{Prathiba2019} \cite{Yao2021} \cite{Deif2020} \cite{Liu2020} \cite{Mohamed2019} \cite{Gupta2020} \cite{Iqbal2020} \cite{Reddy2023} \cite{Pacheco2020} as shown in Table \ref{Table6}.

\subsection{Ablation Study}
\paragraph{} The BCNN has proven highly effective in image classification \cite{Fang2024} \cite{Ying2024} \cite{Cai2020}. This paper introduces a method based on bilinear pooling operation for skin disease classification from the perspective of fine-grained image classification. To assess the performance improvement of the model, we applied compact BCNN with DAM and CCE on two benchmark skin disease image datasets. All the methods used the same experimental steps in the experiment, employing the same data stream except for one variation. The experimental results in Table \ref{Table3} and Table \ref{Table4} indicate that compact BCNN with and without attention enhances the classification accuracy of the model. The results suggest that compact BCNN, with attention, is a notably effective method in medical image classification and well-suited for skin disease classification.

\begin{table*}[!htbp]
\center
\caption{Classification results (Average) of the proposed model on HAM10000.}
\label{Table3}
\renewcommand{\arraystretch}{1.2}%
\begin{tabular}{|*{5}{c|}}
    \hline
\bf{Method} & \bf{Precision} \bf{(\%)}  & \bf{Recall} \bf{(\%)} & \bf{F1-score} \bf{(\%)} & \bf{Accuracy} \bf{(\%)} \\
\hline
ResNet-50 + CCE & 89.80 & 92.78 & 90.61 & 87.69 \\
\hline
Xception + CCE & 89.14 & 89.23 & 89.48 & 89.08 \\
\hline
ResNet-50 + ResNet-50 + CBP + DAM + CCE & 90.08 & 90.45 & 91.54 & 89.74 \\
\hline
Xception + Xception + CBP + DAM + CCE & \bf{91.92} & \bf{93.83} & \bf{94.88} & 91.36 \\
\hline
ResNet-50 + Xception + CBP + CCE & 89.86 & 89.12 & 93.93 & 91.47 \\
\hline
Proposed & 91.72 & 92.34 & 93.03 & \bf{92.86} \\
\hline
\end{tabular}
\end{table*}

\begin{table*}[!htbp]
\center
\caption{Classification results (Average) of the proposed model on ISIC2019.}
\label{Table4}
\renewcommand{\arraystretch}{1.2}%
\begin{tabular}{|*{5}{c|}}
\hline
\bf{Method} & \bf{Precision} \bf{(\%)}  & \bf{Recall} \bf{(\%)} & \bf{F1-score} \bf{(\%)} & \bf{Accuracy} \bf{(\%)} \\
\hline
ResNet-50 + CCE & 88.67 & 87.92 & 92.21 & 87.76 \\
\hline
Xception + CCE & 88.38 & 89.54 & 91.40 & 88.43 \\
\hline
ResNet-50 + ResNet-50 + CBP + DAM + CCE & 91.23 & 89.76 & 92.08 & 88.63 \\
\hline
Xception + Xception + CBP + DAM + CCE & 89.22 & 94.65 & 93.67 & 89.20 \\
\hline
ResNet-50 + Xception + CBP + CCE & \bf{95.64} & 91.36 & 89.11 & 90.14 \\
\hline
Proposed & 92.56 & \bf{99.2} & \bf{95.38} & \bf{94.24} \\
\hline
\end{tabular}
\end{table*}

\begin{table*}[!htbp]
\center
\caption{DACB-Net trained with the different loss functions.}
\label{Table5}
\renewcommand{\arraystretch}{1.2}%
\begin{tabular}{|*{3}{c|}}
\hline
\multirow{2}{*}{\bf{Loss Function}}   & \multicolumn{2}{c|}{\bf{Accuracy} \bf{(\%)}} \\\cline{2-3}
& \bf{HAM10000} & \bf{ISIC2019} \\
\hline
BCE & 91.68 & 91.29 \\
\hline
Focal Loss & 92.62 & 92.77 \\
\hline
CCE & \bf{93.86} & \bf{94.24} \\
\hline
\end{tabular}
\end{table*}

\begin{figure}[!htbp]
\centering
{\includegraphics[scale=0.54]{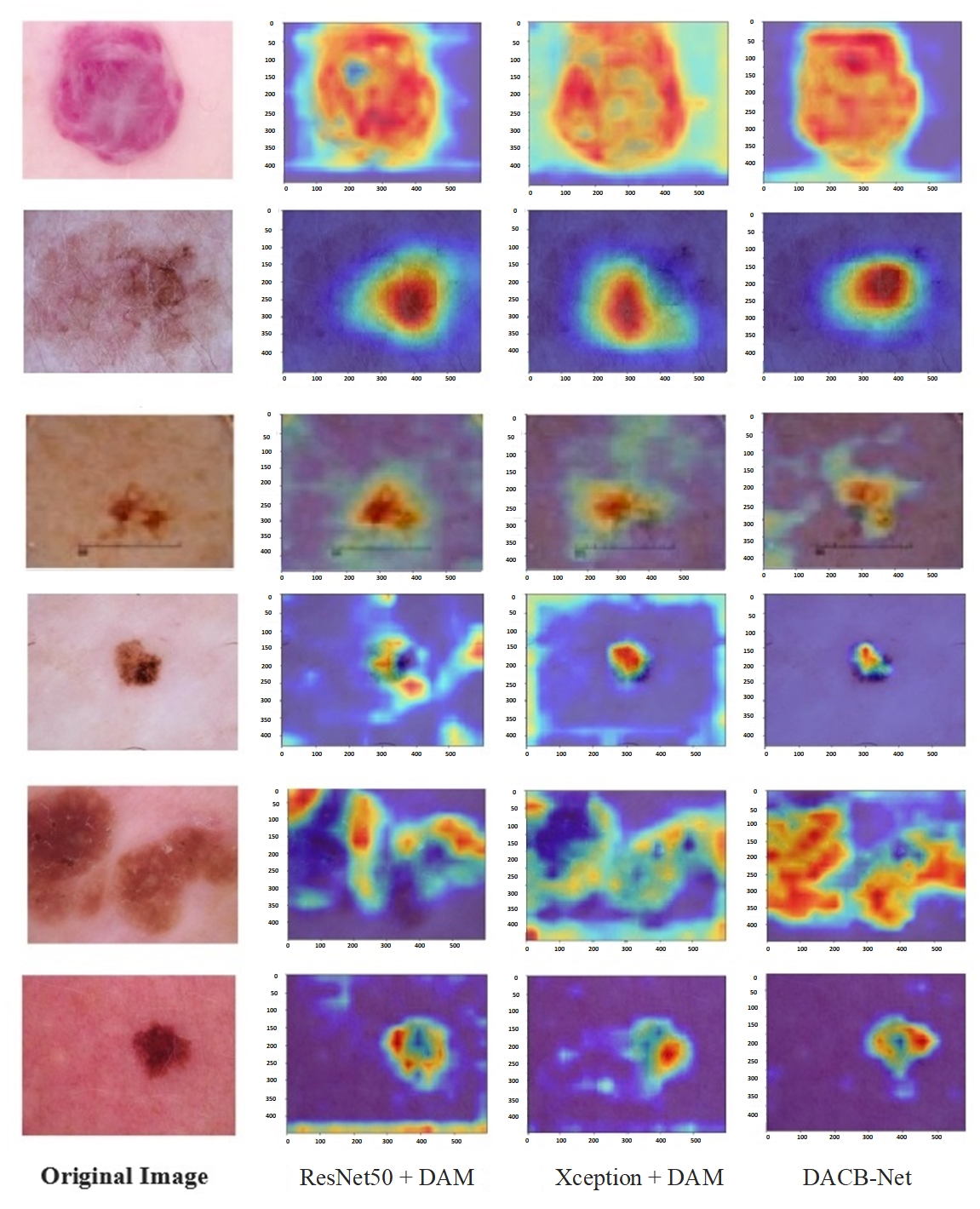}}
\caption{The visualization of the ablation test with the AHM by Grad-CAM on the test set. In the Figure, from left to right, the first column is for the original images, and the second, third, and fourth columns are for ResNet-50 + DAM, Xception + DAM, and Proposed model (DACB-Net), respectively. The higher weight shows a higher thermal value that is classified more correctly.}
\label{Figure8}
\end{figure}

\paragraph{} Following the same experimental steps, further experiments were conducted on the dataset using homologous and non-homologous BCNN. The results show that non-homologous BCNN achieves better classification accuracy than non-homologous BCNN by 1.5\% for HAM10000 and 0.90\% for the ISIC2019 dataset. This suggests that non-homologous BCNN is a superior method in the grading experiment of skin disease images. Non-homologous BCNN concurrently trains two CNNs with different parameter weights, utilizing various data streams to recognize distinct image features. This approach proves more beneficial in extracting specific disease image features.
\paragraph{} We explored by replacing one of the two ResNet-50 networks in the non-homologous BCNN with the Xception. The comparative analysis under a similar experimental setup is detailed in Tables \ref{Table3} and \ref{Table4}. The results indicate that combining different CNNs in BCNN demonstrates their ability to extract more distinctive features from the dataset images, producing better results.
Our investigation compared cross-entropy loss \cite{Deif2020} and focal loss \cite{Deif2020} with CCE loss. All three experiments were performed in the same setup, except for the different loss functions and results outlined in Table \ref{Table5}. The results show that CCE loss surpasses focal loss and BCE in classifying uneven distribution in datasets.
\paragraph{} We evaluated the proposed network on the HAM10000 and ISIC2019 datasets. In addition, these two experiments use the same experimental procedures. It improves the accuracy of the network from 92.86\% to 94.24\%. The performance of the skin disease classification is significantly affected by the attention mechanism. As previously mentioned, the attention mechanism can classify features extracted by the CNN channel by channel, determining dependencies between each channel. Implementing two attention mechanisms enables the network to allocate varying degrees of attention to different image features. The image features enhance the expressive ability of the model to obtain effective results.
\paragraph{} Figure \ref{Figure8} visually represents the attention of the proposed model for different regions in the input image after utilizing the attention mechanism. An original image from the input dataset is shown on the left. In contrast, the images on the right visualize the attention of the network to different regions of the images. In the AHM by Grad-CAM, red denotes a feature in the relevant region (higher weight), while blue indicates the feature’s low weight in the corresponding area.
   
\subsection{Analysis of DACB-Net}
\begin{figure}[!htbp]
\centering
{\includegraphics[scale=0.53]{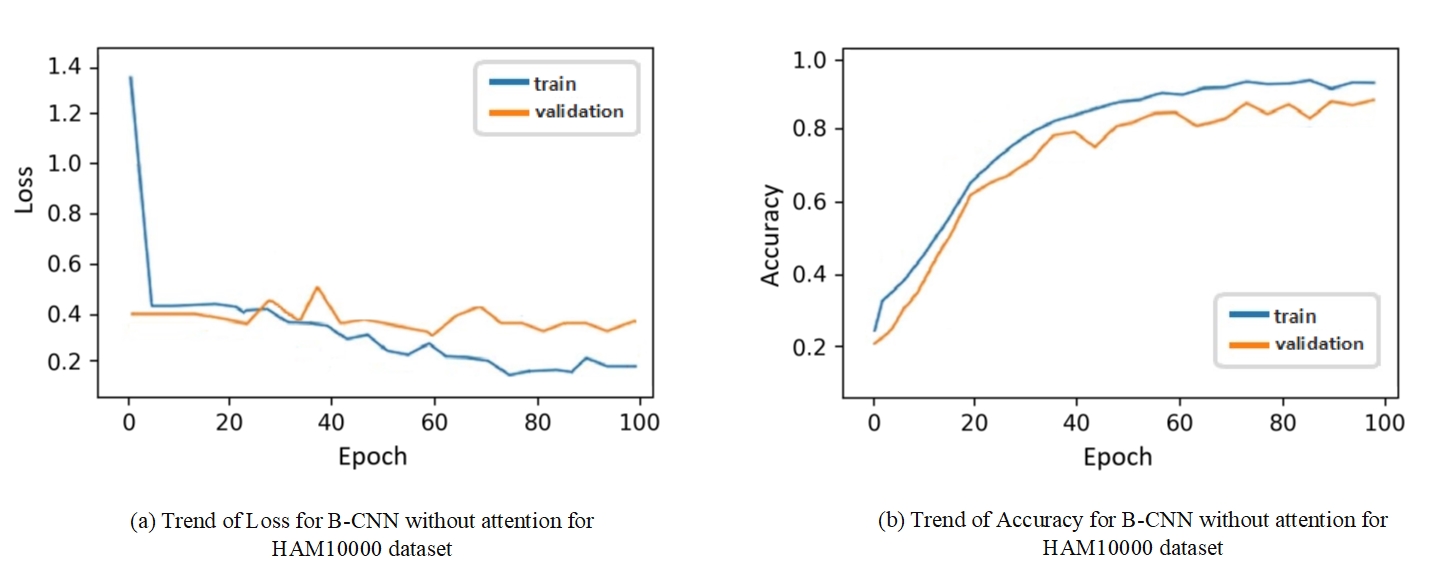}}
\caption{Trend of loss and accuracy for BCNN without attention.}
\label{Figure9}
\end{figure}

\begin{figure}[!htbp]
\centering
{\includegraphics[scale=0.52]{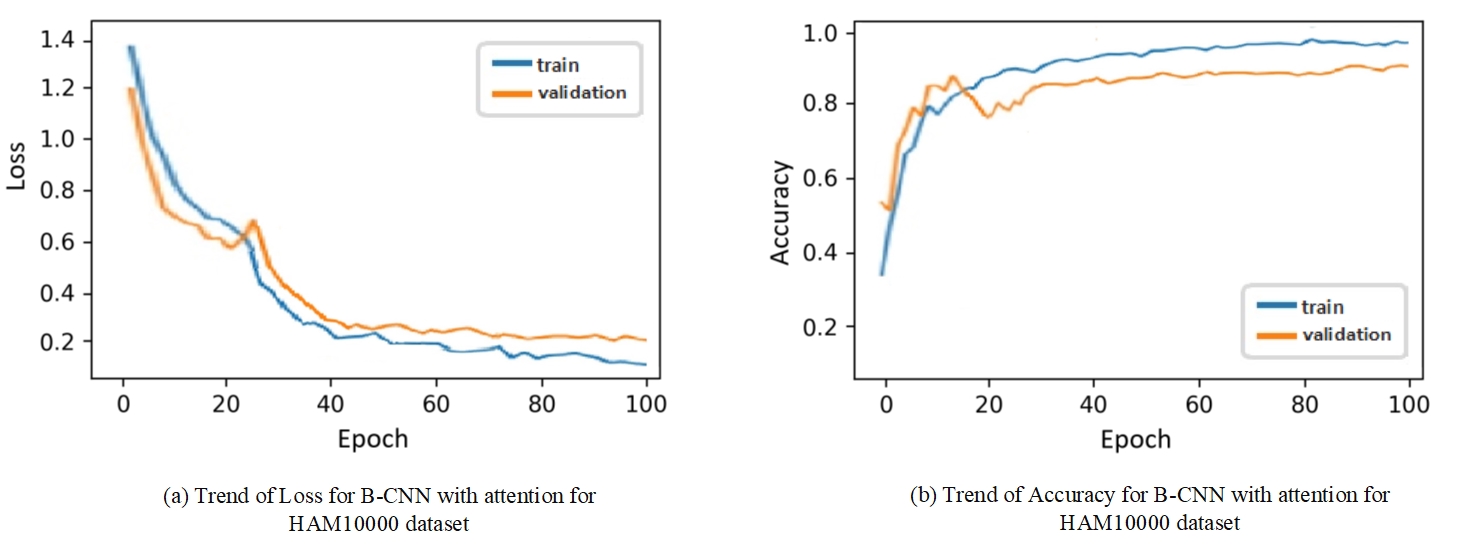}}
\caption{Trend of loss and accuracy for BCNN with attention.}
\label{Figure10}
\end{figure}

\begin{figure}[!htbp]
\centering
{\includegraphics[scale=0.53]{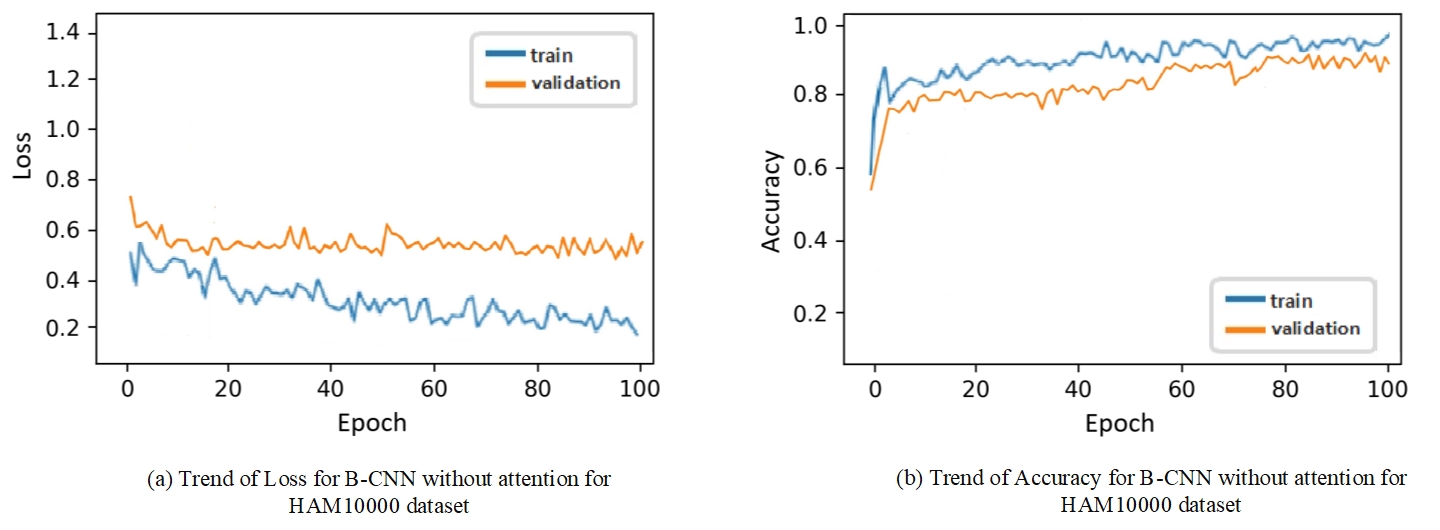}}
\caption{Trend of loss and accuracy for DACB-Net for HAM10000.}
\label{Figure11}
\end{figure}

\begin{figure}[!htbp]
\centering
{\includegraphics[scale=0.53]{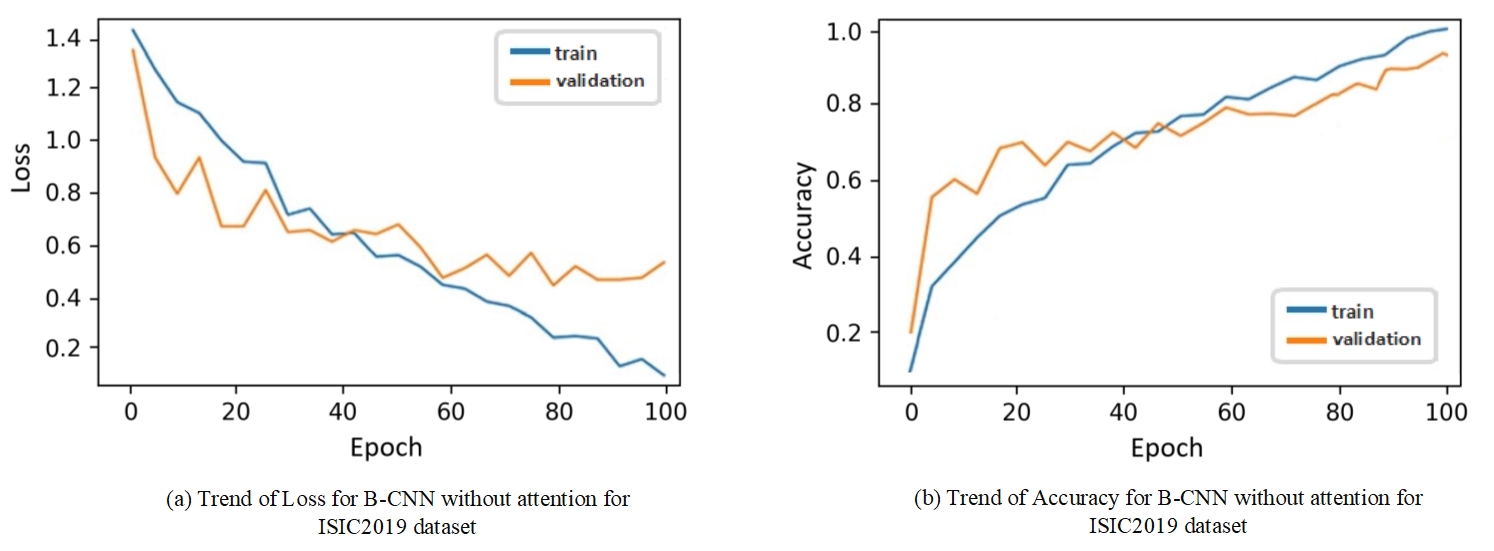}}
\caption{Trend of loss and accuracy for BCNN without attention on the ISIC2019 dataset.}
\label{Figure12}
\end{figure}

\begin{figure}[!htbp]
\centering
{\includegraphics[scale=0.55]{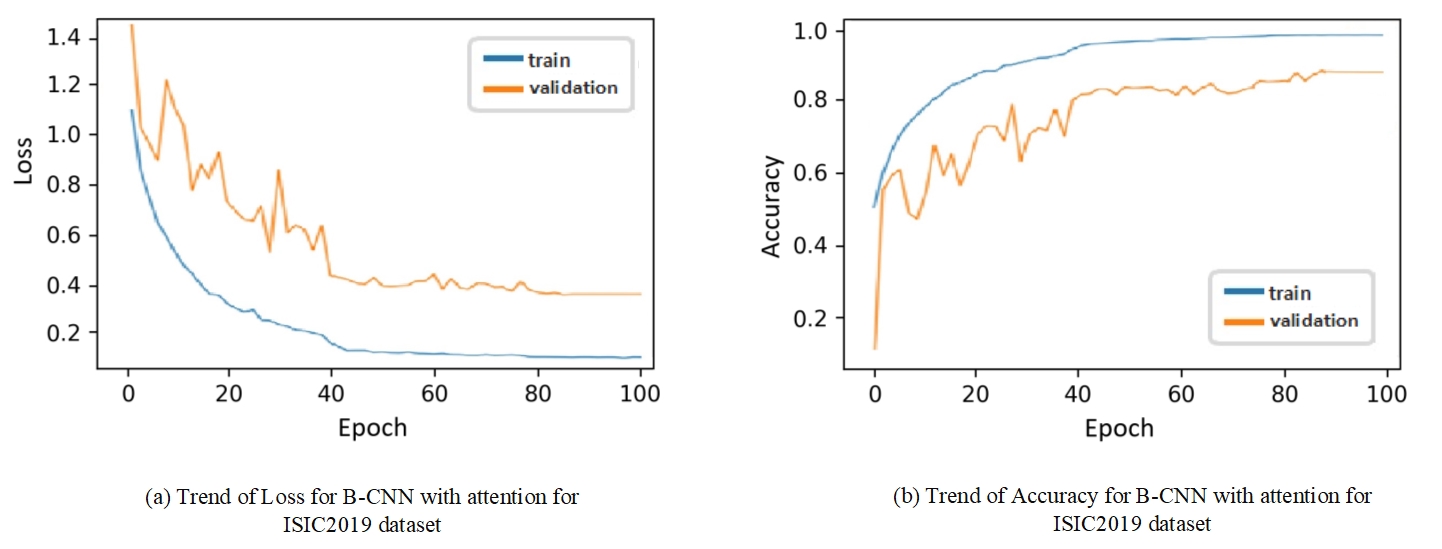}}
\caption{Trend of loss and accuracy for BCNN with attention.}
\label{Figure13}
\end{figure}

\begin{figure}[!htbp]
\centering
{\includegraphics[scale=0.53]{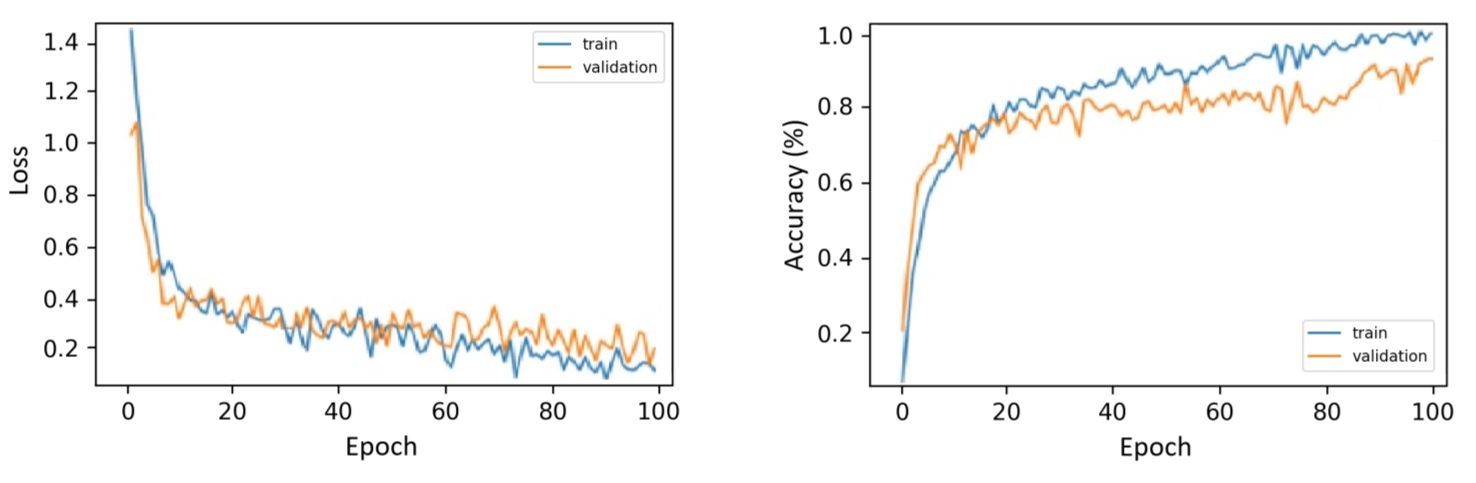}}
\caption{Trend of loss and accuracy for DACB-Net for ISIC2019.}
\label{Figure14}
\end{figure}

\paragraph{} The proposed model was trained and evaluated with five folds to check its effectiveness for classification. The architecture details are described in the above section. The model was trained with an Adam optimizer for 50 epochs on each fold. The accuracy and loss curves for the training and validation data are shown in Figures \ref{Figure9} - \ref{Figure14}.
\paragraph{} Although the training accuracy increased gradually during the training, it started initially faster and later slowed significantly. With the training, the validation accuracy fluctuated while it was increasing. The validation curves of the proposed model are relatively less staggered than the compact bilinear model without an attention mechanism.
\paragraph{} The validation curve is relatively less staged than the compact BCNN without DAM. It reaches the saturation points at around 50 epochs. The margin between training and validation curves was minimized compared to the compact BCNN without DAM. Using the validation data, the proposed model achieved a validation accuracy of 94.24\%, with a variance of ± 3.4\% over the five folds. This confirms that when test data is unknown, the model fitting performs better at generalizing than compact BCNN without DAM.

\subsection{Performance Analysis}
\begin{figure}[!htbp]
\centering
{\includegraphics[scale=0.57]{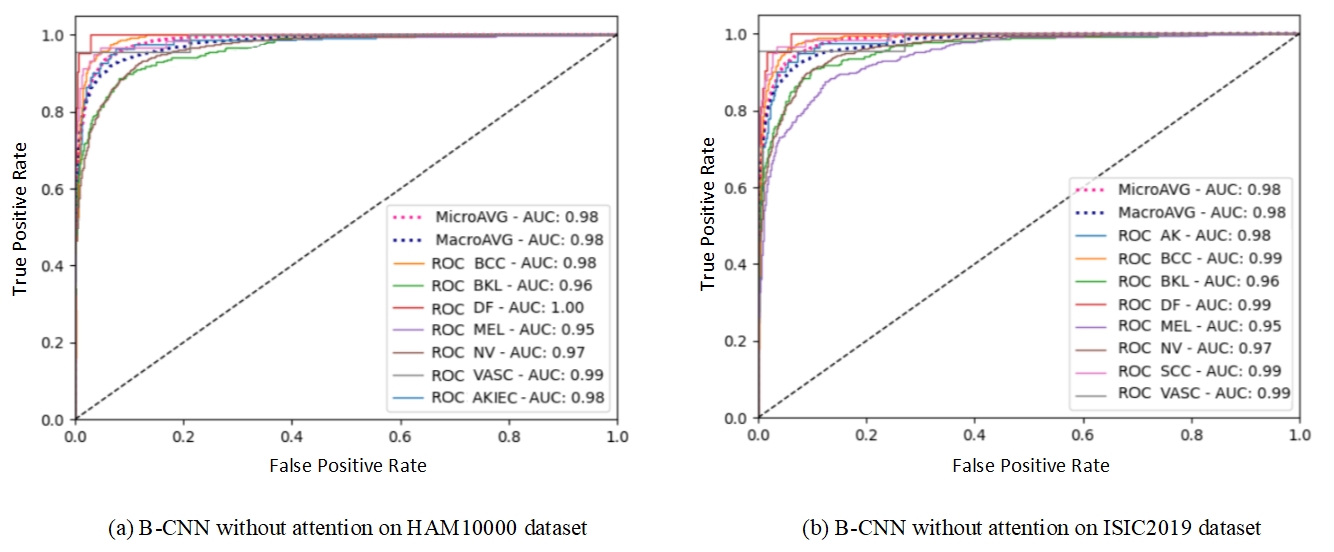}}
\caption{ROC Curve of ResNet-50 and Xception without attention for (a) HAM10000 and (b) ISIC2019.}
\label{Figure15}
\end{figure}

\begin{figure}[!htbp]
\centering
{\includegraphics[scale=0.60]{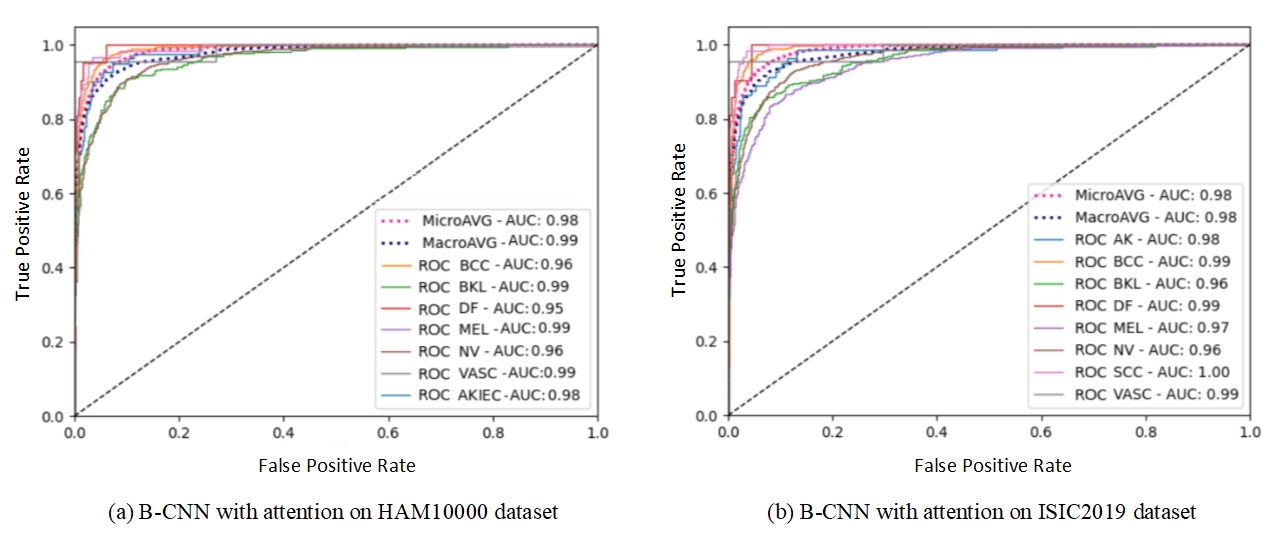}}
\caption{ROC Curve of ResNet-50 and Xception with attention for (a) HAM10000 and (b) ISIC2019.}
\label{Figure16}
\end{figure}

\begin{figure}[!htbp]
\centering
{\includegraphics[scale=0.61]{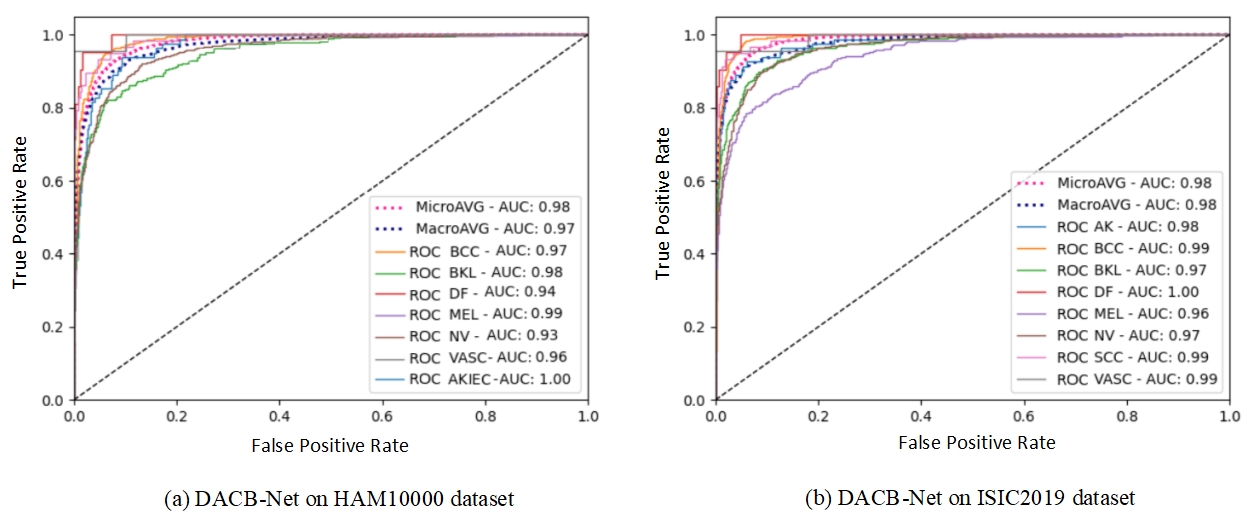}}
\caption{ROC Curve of DACB-Net for (a) HAM10000 and (b) ISIC2019 dataset.}
\label{Figure17}
\end{figure}

\begin{figure}[!htbp]
\centering
{\includegraphics[scale=0.38]{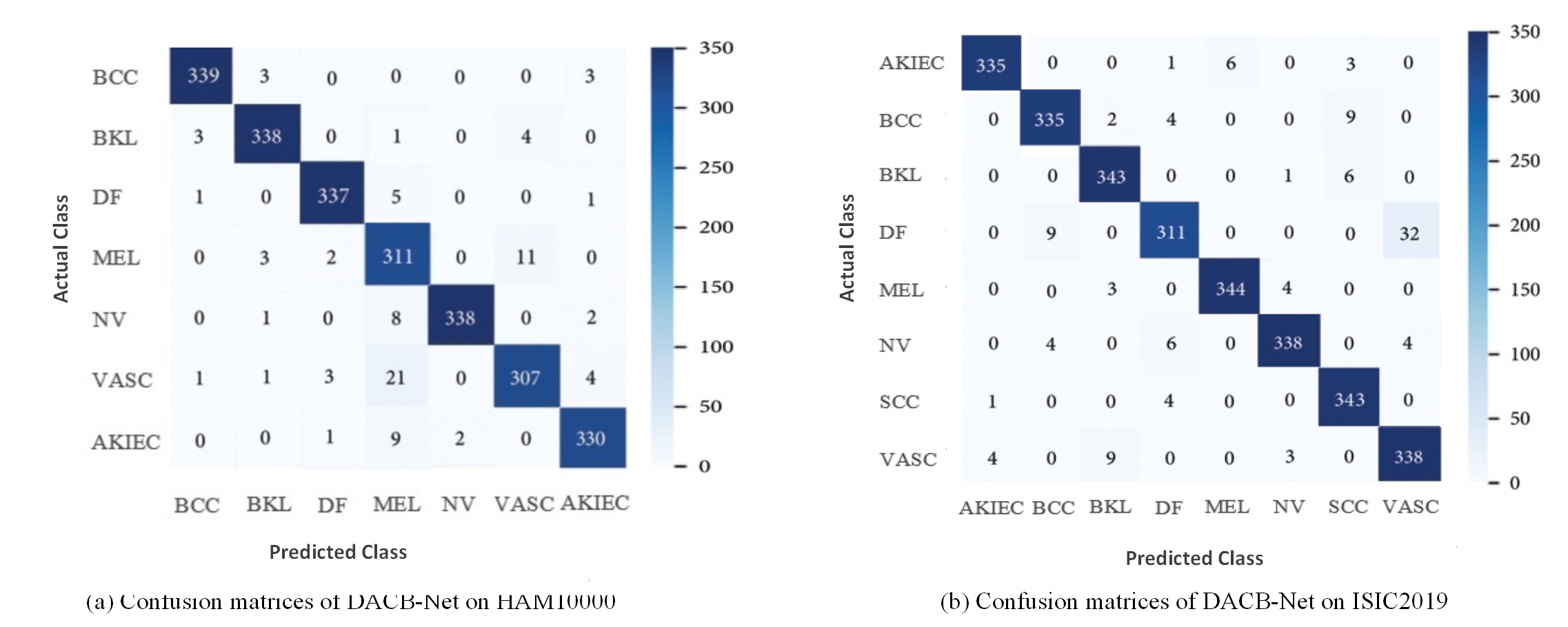}}
\caption{Confusion matrices of the proposed model on (a) HAM10000 and (b) ISIC2019.}
\label{Figure18}
\end{figure}

\paragraph{} The ROC curves depicting the classification of all classes in the HAM10000 and ISIC2019 datasets are shown in Figure \ref{Figure15} - \ref{Figure17}. The AUC values corresponding to each curve are indicated in the labels, along with micro and macro averages. The micro-average considers contributions from all classes to compute the average metric, while the macro-average calculates the metric independently for each category and then determines the average. Notably, in both bilinear methods, the extreme points of curvature in the graphs are positioned close to the upper left corner, suggesting a good separation capacity of the evaluated model. Substituting the ECA block in the proposed architecture significantly reduced model complexity and trainable parameters. The proposed model achieved 92.86 \% and 94.24\% accuracy on the HAM10000 and ISIC2019 datasets, respectively.
\paragraph{} The confusion matrix, a quantitative-visual representation of classification results, is displayed for the proposed model in Figure \ref{Figure18} for (a) HAM10000 and (b) ISIC2019 datasets. The vertical axis represents true labels in these matrices, and the horizontal axis denotes predicted labels. Ideally, identity matrices are expected, indicating the perfect classification of all images. Errors in classification are reflected outside the main diagonal. For instance, in Figure 18 (a), a value of 339 was obtained for the "BCC" class due to some images being predicted as "BKL". Similar confusion between the "AKIEC" and "MEL" classes is evident in Figure \ref{Figure18} (b). Additionally, there were labelling errors for "BKL", predicted as "BCC" and "BKL". Despite these errors, both confusion matrices indicate highly accurate results.

\begin{table*}[!htbp]
\center
\caption{Comparison with state-of-the-art methods.}
\label{Table6}
\renewcommand{\arraystretch}{1.45}%
\begin{tabular}{|*{6}{c|}}
\hline
\bf{Method} & \bf{Precision \bf{(\%)}} & \bf{Sensitivity \bf{(\%)}} & \bf{Specificity \bf{(\%)}} & \bf{F1-score \bf{(\%)}} & \bf{Accuracy \bf{(\%)}} \\ [1mm]
\hline
MobileNetV2-LSTM \cite{Srinivasu2021} & - & 88.24 & 92.00 & - & 85.34 \\
\hline
Kassem et al. \cite{Kassem2020} & - & 74.00 & 84.00 & - & 81.00 \\
\hline
Raju et al. \cite{Raju2022} & 91.18 & 92.20 & 91.92 & 92.28 & 92.81 \\
\hline
Bharat et al. \cite{Bharat2022} & 94.46 & 94.92 & 94.83 & 93.13 & 93.95 \\
\hline
Kumar et al. \cite{Kumar2022} & 95.72 & 95.71 & 95.21 & 95.25 & 94.19 \\
\hline
Brinker et al. \cite{Brinker2022} & - & 88.88 & 91.77 & - & 90.33 \\
\hline
MKhan et al.\cite{Khan2018} & - & 88.50 & 91.00 & - & 88.20 \\
\hline
MobileNetV2 \cite{Indraswari2022} & - & 85.00 & 85.00 & - & 85.00 \\
\hline
Prathiba et al. \cite{Prathiba2019} & - & 85.30 & \textbf{99.30} & - & 85.50 \\
\hline
MWNL-CLS \cite{Yao2021} & - & 56.40 & 76.00 & - & 76.30 \\
\hline
InceptionV3 \cite{Deif2020} & 88.00 & 75.00 & - & 81.00 & 89.81 \\
\hline
Semi-supervised \cite{Liu2020} & 88.00 & 71.47 & - & 60.68 & 92.54 \\
\hline
MobileNet \cite{Mohamed2019} & 87.00 & 81.00 & - & 84.00 & 92.70 \\
\hline
EfficientNetB1\cite{Gupta2020} & 94.00 & 94.00 & - & 94.00 & 94.00 \\
\hline
CSLNet \cite{Iqbal2020} & 92.56 & \bf{99.20} & 97.57 & 89.75 & 89.58 \\
\hline
Reddy et al. \cite{Reddy2023} & \bf{96.15} & 91.83 & 96.47 & 93.94 & 94.20 \\
\hline
DBN Small \cite{Pacheco2020} & 85.89 & 84.20 & 97.58 & 85.03 & 91.58 \\
\hline
Proposed & 94.56 & \bf{99.20} & - & \bf{95.38} & \bf{94.24} \\
\hline
\end{tabular}
\end{table*}

\subsection{Comparison}
\paragraph{} To examine the concern of the proposed model, we compared the results of the proposed method with some existing methods for skin disease classification. Performance comparison metrics included precision, recall, F1-score, and model accuracy. Table \ref{Table6} presents the resulting observations, and the accuracy and F1-score of the proposed method achieved the best results compared to the existing methods, whereas the precision of \cite{Reddy2023}, and specificity of \cite{Prathiba2019} is more than the DACB-Net, but their sensitivity and accuracy is comparatively very low. So, it was concluded that the proposed DACB-Net achieved the best results compared to the existing methods.

\section{Conclusion and Future Work}
\paragraph{} In this paper, we address the skin disease image classification problem. The final classification accuracy of the proposed network reached a state-of-the-art level. In this network, we used a compact bilinear attention mechanism by improving the CAM and SAM design and restructuring with the bilinear concept, which is more effective in extracting discriminative features. Furthermore, the network employed two supervision branches along with a new loss function. This approach calculated gradients from distinct pairs of local image patches and incorporated a modulation factor in the loss function to prioritize training on complex samples. It handles the data imbalance problem by adjusting the weight, making it vulnerable to overfitting and enhancing the proposed network’s performance and interpretability. Additionally, using the ECA block instead of the SE block in the proposed network reduces the model complexity and trainable parameters. BCNN is high-dimensional and affects the computational cost and memory use for further feature analysis. To address this, we used a compact bilinear pooling network to perform core analysis on bilinear features, significantly reducing dimensionality. This operation allows the proposed model to optimize memory usage while maintaining significant classification accuracy.
\paragraph{} Furthermore, we visually analyzed the AHM of the two mechanisms using Grad-CAM to evaluate the classification accuracy. Second, we conducted an ablation test to verify the importance of each component in the attention framework. The effectiveness of the framework structure in enhancing the network feature extraction ability was confirmed. Finally, we evaluated the proposed network’s superiority and robustness by comparing its results with the state-of-the-art methods for skin disease classification. The results demonstrated that the proposed network exhibits higher accuracy and better robustness in implementing skin disease classification.
\paragraph{} In the future, DACB-Net may be combined or used with other network structures for other tasks to show the generality. However, due to the limitations of hardware resources, it has not been tested with different specifications. The research results of the proposed network show that its accuracy increases continuously with the expansion of specification. Therefore, theoretically, if other network specifications are used, the accuracy and their degree of optimization can be further improved. Furthermore, since the proposed algorithm and the optimization approach are experimentally validated on the existing dataset, the model can be subsequently used to accomplish fine-grained classification tasks in practical applications.

\bibliographystyle{unsrt}  
\bibliography{Manuscript}  

\end{document}